\documentclass[
]{revtex4-2}

\usepackage{booktabs}
\usepackage{graphicx}
\usepackage{dcolumn}
\usepackage{bm}
\usepackage{amsmath}
\usepackage{bbm}
\usepackage{bm}
\usepackage{xcolor}
\usepackage{makecell}
\usepackage{cellspace} 
\usepackage{soul}
\usepackage{amssymb}
\sethlcolor{white}
\usepackage{lineno}

\def\tsc#1{\csdef{#1}{\textsc{\lowercase{#1}}\xspace}}
\tsc{WGM}
\tsc{QE}

\usepackage{comment}

\begin{document}
\title{KAN-ODEs: Kolmogorov-Arnold Network Ordinary Differential Equations \\ for Learning Dynamical Systems and Hidden Physics }

\author{Benjamin C. Koenig}
\thanks{B.C.K. and S.K. contributed equally to this work.}
\affiliation{Department of Mechanical Engineering, Massachusetts Institute of Technology, 77 Massachusetts Ave, Cambridge, MA 02139, United States.}

\author{Suyong Kim}
\thanks{B.C.K. and S.K. contributed equally to this work.}
\affiliation{Department of Mechanical Engineering, Massachusetts Institute of Technology, 77 Massachusetts Ave, Cambridge, MA 02139, United States.}

\author{Sili Deng}
\email{Corresponding author, silideng@mit.edu}
\affiliation{Department of Mechanical Engineering, Massachusetts Institute of Technology, 77 Massachusetts Ave, Cambridge, MA 02139, United States.}
\date{\today}

\begin{abstract}
Kolmogorov-Arnold networks (KANs) as an alternative to multi-layer perceptrons (MLPs) are a recent development demonstrating strong potential for data-driven modeling. This work applies KANs as the backbone of a neural ordinary differential equation (ODE) framework, generalizing their use to the time-dependent and temporal grid-sensitive cases often seen in dynamical systems and scientific machine learning applications. The proposed KAN-ODEs retain the flexible dynamical system modeling framework of Neural ODEs while leveraging the many benefits of KANs compared to MLPs, including higher accuracy and faster neural scaling, stronger interpretability and generalizability, and lower parameter counts. First, we quantitatively demonstrated these improvements in a comprehensive study of the classical Lotka-Volterra predator-prey model. We then showcased the KAN-ODE framework's ability to learn symbolic source terms and complete solution profiles in higher-complexity and data-lean scenarios including wave propagation and shock formation, the complex Schr\"{o}dinger equation, and the Allen-Cahn phase separation equation. The successful training of KAN-ODEs, and their improved performance compared to traditional Neural ODEs, implies significant potential in leveraging this novel network architecture in myriad scientific machine learning applications for discovering hidden physics and predicting dynamic evolution.

\end{abstract}

\maketitle

\section{\label{sec:intro}Introduction}
Dynamical system modeling is a key part of many branches of engineering and science. Traditionally, governing equations (often partial differential equations (PDEs)) have been derived through extensive observations and careful selection of model equations. These processes, heavily relying on expert knowledge, have gradually been augmented and sometimes replaced with machine learning techniques by leveraging the data-fitting power in neural networks \cite{karniadakis2021physics,brunton2016discovering,brunton2024promising,rudy_data-driven_2017}. Despite the potential to reduce the need for detailed expert knowledge, these data-driven modeling approaches are often subject to tradeoffs between the amount of prior knowledge embedded in the model versus the interpretability of the model: purely data-driven approaches tend to lack interpretability, and conversely, interpretable methods typically require substantial knowledge or enforcement of preexisting governing laws \cite{karniadakis2021physics}. To mitigate this issue, there have been extensive efforts to develop novel modeling schemes that blend data-driven modeling with traditional dynamical system tools, including sparse identification of nonlinear dynamics (SINDy) \cite{brunton_discovering_2016}, physics-informed neural network (PINNs) \cite{raissi_physics-informed_2019}, and neural ordinary differential equations (Neural ODEs) \cite{chen_neural_2019}. Such studies that bridge this gap have the potential to not only infer high-accuracy models from data but also provide useful insight into the dynamics of what are often unknown physical processes.

Neural ODEs have been developed to learn a continuous evolution of states \cite{chen_neural_2019, kim_stiff_2021, dandekar_bayesian_2022}, as opposed to the discrete sequences learned by residual networks \cite{lu2018beyond,haber2017stable}. Chen et al. coupled black box neural networks used as gradient getters with standard ODE solvers in order to learn the relationship between the current state of a system and its gradient \cite{chen_neural_2019}. This mapping gives Neural ODEs inherent grid and timescale flexibility, allowing for substantial interpolation and generalization of learned models compared with typical fixed grid neural network approaches. Neural ODEs do not require prior knowledge of a system in order to infer dynamical models from large datasets thanks to their use of black box multi-layer perceptrons (MLPs). However, on the flip side of this benefit, they typically deliver models with thousands to millions of parameters, making any deeper interpretation or extraction of physical insights difficult.

In contrast to Neural ODEs, physics-informed neural networks (PINNs) leverage the governing equations of the dynamical system, wherever available, in the loss function to train a traditional (i.e. not ODE based) neural network that maps the time and spatial coordinates directly to the state variables \cite{raissi_physics-informed_2019, ji_stiff-pinn_2021, cuomo_scientific_2022}. This physics-embedded loss function allows for training that aims to fit both the data and the known governing equations without having to directly solve what are typically expensive PDEs. These networks are significantly interpretable thanks to the governing equations used as a soft constraint in the training cycle, giving physical meaning to trends and patterns that emerge from the training process. The inclusion of the governing equations as a soft constraint also allows PINNs to train with relatively small datasets, which is a useful benefit in many scientific and engineering applications with high costs of data acquisition \cite{raissi_physics-informed_2019}. Other frameworks have also been developed using a physics-based structure similar to PINN. The auto-regressive dense encoder-decoder model \cite{geneva_modeling_2020} encodes physics into a time-integration network scheme, while the physics-informed convolutional-recurrent network \cite{ren_phycrnet_2022} runs PDE solutions through convolutional filters with physical penalty terms and hard-encoded initial and boundary conditions to efficiently develop models. Other works have also investigated the use of hard constraints with PINN-type models for initial and boundary conditions with strong results \cite{lu_physics-informed_2021, alkhadhr_wave_2023}. PINN-type loss functions have additionally been included in Neural ODE frameworks \cite{lai_structural_2021}. Further combinations exist in the area of physics-based Neural ODEs, such as Chemical Reaction Neural Networks \cite{ji_autonomous_2021}, which hard constrain Neural ODE frameworks to directly learn governing equation parameters from real \cite{ji_autonomous_2022, koenig_accommodating_2023} and noisy \cite{li_bayesian_2023, koenig_uncertain_2024} experimental data. A downside shared throughout these numerous physics-based techniques is their requirement of strong prior knowledge of the system's governing laws. In cases where such knowledge is not available, other methods are required. 

The SINDy approach \cite{brunton_discovering_2016, kaiser_sparse_2018, fasel_ensemble-sindy_2022, rudy_data-driven_2017} tackles dynamical system modeling via sparse regression. Given training data and a set of candidate functions (polynomials, trigononometric functions, constant functions, etc.), it carries out sparse regression to parsimoniously select only the candidate functions needed to describe trends present in the data. The use of a presumed set of candidate functions gives SINDy models significant interpretability over pure black box methods like MLPs or Neural ODEs. It also requires less of the detailed physical knowledge behind the system dynamics as in PINNs and other physics-based methods, and less data with a reduced model search space compared to Neural ODEs.

While most advances in data-driven modeling lie in MLPs and polynomial basis approaches, Liu et al. recently proposed Kolmogorov-Arnold networks (KANs) as an alternative to MLP-based networks \cite{liu_kan_2024}. Instead of learning weights and biases given fixed activation functions as in MLPs, the activation functions themselves are learnable in KANs with gridded basis functions and associated trainable scaling factors. Major advantages of KANs over MLPs are increased interpretability thanks to the direct learning of activation functions, faster convergence to low training loss with significantly fewer parameters thanks to their faster neural scaling laws, and the possibility of postprocessing via sparse regression to deliver human-readable expressions. With this potential, recent efforts have been dedicated to extending KANs for time-series predictions based on residual networks by directly mapping training data onto future testing windows \cite{vaca-rubio_kolmogorov-arnold_2024, xu_kolmogorov-arnold_2024}. However, such a discrete time-step approach has a limited ability to predict the states where various time scales and data collection windows are involved. This research gap naturally motivates our work to embed KANs into Neural ODE-type dynamical system solvers, allowing us to resolve these drawbacks while preserving the interpretability and inference potential of KANs.

\begin{figure}[tb]
\centering
	\includegraphics[width=0.4\linewidth]{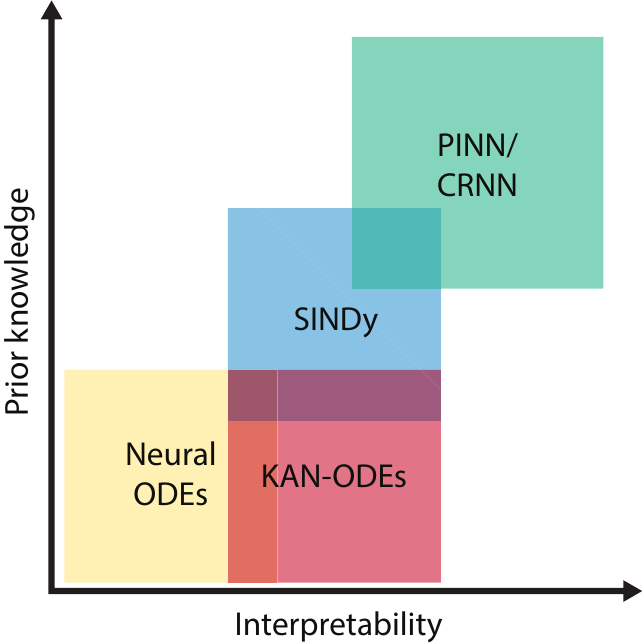}
	\caption{Qualitative depiction of KAN-ODE's general capability compared to similar machine learning techniques.}
	\label{fig:intro}
\end{figure}

In this study, we introduce the possibility of learning interpretable and modular ODE and PDE models for dynamical systems without prior knowledge or functional form assumptions by leveraging KANs as gradient-getters within the concept of Neural ODEs. In doing so, we propose a combined framework that benefits from the high-accuracy, parameter-lean, and strongly interpretable KAN network structure coupled with the general-purpose, grid-independent, and solver-flexible Neural ODE framework. A qualitative depiction of the comparison between the currently proposed KAN-ODE method and the other discussed data-driven approaches for dynamical system modeling is shown in Fig.~\ref{fig:intro}. While the current state of the art is generally situated on the diagonal of this plot, where increased prior knowledge fed into a machine learning method delivers better interpretability, we place KAN-ODEs slightly below the diagonal, thanks to their potential to deliver interpretable activations and even symbolic relationships without requiring prior knowledge or presumed functional forms in the training cycle. The ODE-based formulation is also inherently flexible, allowing for variable timesteps depending on the gradient landscape as well as the application of the KAN-ODE as a component of a larger-scale model, for example as the source term in a nonlinear transport PDE. We demonstrated the promise of KAN-ODEs through a comprehensive comparison against standard Neural ODEs, as well as an array of tests at different scales and levels of complexity including a predator-prey ODE, wave propagation and shock formation PDEs, the complex Schr\"{o}dinger equation, and a phase separation PDE. We found that KAN-ODEs succeed in all of these cases and are able to reconstruct complete temporal profiles even when given sparse training data. KAN-ODEs with a small number of parameters were able to converge with remarkable accuracy, and sparsity or symbolic relationships imposed during the training cycle or during postprocessing were able to give KAN-ODEs substantial generalization capability. Overall, we found that this encoding of KANs into the Neural ODE framework is a promising step toward small-data and low-knowledge yet interpretable machine learning for dynamical system modeling.

\section{\label{sec:methods_overview}Methodology}

The formulation proposed here leverages KANs and the neural network-ODE solver concept of Neural ODEs. Sec.~\ref{sec:methods_kan} introduces the concept of a KAN. Then, we describe how we couple KANs to differentiable ODE solvers to model dynamical systems in Sec.~\ref{sec:methods_NODE}.

\subsection{\label{sec:methods_kan}Kolmogorov Arnold Networks as Gradient Evaluators}

In contrast to MLPs based on the universal approximation theorem \cite{hornik_multilayer_1989}, KANs \cite{liu_kan_2024} are based on the Kolmogorov-Arnold representation theorem (KAT) \cite{kolmogorov_representation_1956}, which states that any multivariate function $f(\textbf{x}) : [0, 1]^n \rightarrow \mathbb{R}$ that is continuous and smooth can be represented by a finite sum of continuous univariate functions, 

\begin{equation} \label{eq:KAN}
f(\textbf{x})=f(x_1, ..., x_n) = \sum_{q=1}^{2n+1} \Phi_q \left( \sum_{p=1}^n \phi_{q, p}(x_p) \right).
\end{equation}

\noindent Here, $\phi_{q, p}:~[0, 1] \rightarrow \mathbb{R}$ are univariate functions, while $\Phi_{q} : \mathbb{R} \rightarrow \mathbb{R}$ composes these univariate functions to reconstruct $f(\textbf{x})$. The implementation in this work defines $\Phi_{q}$ as trainable weights to sum the univariate basis functions. Once these basis functions are chosen, Eq.~\ref{eq:KAN} exactly describes a 2-layer KAN with a hidden layer width of $2n+1$ (the upper bound of the outer summation in Eq.~\ref{eq:KAN}) and an output dimension of 1. The KAT is extended to a general KAN structure with $L$ layers of arbitrary width by exploiting the hierarchical structure of Eq.~\ref{eq:KAN} \cite{liu_kan_2024}.

\begin{equation}\label{eq:KAN_forward}
    \mathbf{y}={\text{KAN}}(\mathbf{x}) = \left({\Phi}_{L-1}\circ {\Phi}_{L-2}\circ\cdots\circ {\Phi}_{1}\circ {\Phi}_{0}\right)\left(\mathbf{x}\right).
\end{equation}

\begin{equation}\label{eq:kanforwardmatrix}
    {\Phi}_l = 
    \begin{pmatrix}
        \phi_{l,1,1}(\cdot) & \phi_{l,1,2}(\cdot) & \cdots & \phi_{l,1,n_{l}}(\cdot) \\
        \phi_{l,2,1}(\cdot) & \phi_{l,2,2}(\cdot) & \cdots & \phi_{l,2,n_{l}}(\cdot) \\
        \vdots & \vdots & & \vdots \\
        \phi_{l,n_{l+1},1}(\cdot) & \phi_{l,n_{l+1},2}(\cdot) & \cdots & \phi_{l,n_{l+1},n_{l}}(\cdot) \\
    \end{pmatrix},
\end{equation}

\noindent where $\phi$ is the activation function, $l$ is the index of a layer, and $n_{l}$ and $n_{l+1}$ denote the numbers of nodes in the $l^{\text{th}}$ and $(l+1)^{\text{th}}$ layers, respectively. While Liu et al. \cite{liu_kan_2024} originally proposed a B-spline for the gridded basis functions that comprise the various activations $\phi$, this study adopts the more computationally efficient Gaussian radial basis functions (RBFs) $\psi$ as proposed by Li \cite{li_kolmogorov-arnold_2024}, such that

\begin{align}
    \phi_{l, \alpha, \beta} \left(\text{x} \right) &= \sum_{i=1}^{N} w^{\psi}_{l, \alpha, \beta,i} \cdot \psi \left( \lvert \lvert \text{x}-c_{i} \rvert \rvert \right) + w^b_{l, \alpha, \beta} \cdot b\left(\text{x}\right),\label{eq:basis}\\
    \psi(r)&=\exp(-\frac{r^{2}}{2h^{2}}), \label{eq:RBF}
\end{align}

\noindent where $N$ is the number of gridpoints, $w^{\psi}_{l, \alpha, \beta,i}$ and $w^b_{l, \alpha, \beta}$ are the trainable weights for the RBF function $\psi(r)$ and the base activation function $b(\text{x})$ respectively (superscripts indicating which function the weights apply to), the subscripts $[\alpha, \beta]$ indicate the index of the matrix in Eq. \ref{eq:kanforwardmatrix} within a KAN layer $l$ that these weights apply to, $c_{i}$ is the center of the $i$-th RBF function, $r$ is the distance of $\text{x}$ from the center, and $h$ is a spreading parameter defined as the gridpoint spacing.  The inputs to each layer are normalized to be on the [-1, 1] gridded RBF domain as in \cite{puri_kolmogorovarnoldjl_2024}, which avoids the costly alternative technique of \cite{liu_kan_2024} that requires periodically re-gridding the RBF networks in each layer to match the input domain. In addition to fully gridded RBF basis functions, the inputs and outputs of each layer are also directly connected with Swish residual activation functions such that $b(\text{x})=\text{x}\cdot \text{sigmoid}(\text{x})$ \cite{ramachandran_searching_2017}.

Equation \ref{eq:basis} therefore defines two input pathways for each layer: the gridded RBF pathway (first term), and the direct (non-gridded) residual activation pathway (second term). The RBF function connects each input node with each output node using a number of basis functions equal to the size of the grid. The number of parameters required for the RBF pathway of a single layer is therefore equal to the product of the input dimension, output dimension, and grid size (following as well from $w^{\psi}_{l, \alpha, \beta,i}$, where for a single layer $w^{\psi}_{l}$ the remaining three indices cover the input, output, and grid indices). Similarly, the residual activation pathway, which directly connects inputs to outputs without a grid, simply requires parameters equal to the product of the input and output dimensions (likewise see $w^b_{l, \alpha, \beta}$, which includes the input and output indices $\alpha$ and $\beta$ but not the grid index $i$). For example, in a 2-layer KAN with a 2D input, 1D output, 10D hidden layer, and 5-point grid, the numbers of parameters are 2$\times$5$\times$10+2$\times$10 for the input $\rightarrow$ hidden layer, and 10$\times$5$\times$1+10$\times$1 for the hidden $\rightarrow$ output layer, respectively, resulting in 180 parameters in total.

Expanded mathematical formulations for KANs with multivariate outputs become increasingly bulky and cumbersome and are available in the extensive \cite{liu_kan_2024}. For convenience, we use the notation [$n_{l}, n_{l+1}, N$] to represent an $l$-th KAN layer for the remainder of this work, where $n_{l}$ is the input dimension, $n_{l+1}$ is the output dimension, and $N$ is the grid size. For example, a 2-input and 1-output KAN with a 10-node hidden layer and a 5-point grid is represented as "[2, 10, 5], [10, 1, 5]" for the connections between the input and hidden layer, and the hidden and output layer, respectively.

\subsection{\label{sec:methods_NODE}KAN-Ordinary Differential Equations}

A dynamical system is generally expressed by an ODE system such that

\begin{equation}
    \frac{d\mathbf{u}}{dt} = \mathbf{g}\left(\mathbf{u},t \right),
\end{equation}

\noindent where $\mathbf{u} \in \mathbb{R}^{n}$ is a vector of the state variables and $\mathbf{g}$ is the system equation. Chen et al. \cite{chen_neural_2019} previously proposed neural ordinary differential equations with the help of the universal approximation theorem (UAT) and the \textit{adjoint} sensitivity method to construct a model $\text{NN}$ such that $ \text{NN} \approx \mathbf{g}$. We similarly adopt the recently developed KAN as an approximator for a dynamical system by leveraging the benefits of the Kolmogorov-Arnold representation theorem (KAT). The proposed KAN-ODEs can be expressed as

\begin{equation} \label{eq:KAN_gradient}
\frac{d\mathbf{u}}{dt}=\text{KAN}\left(\mathbf{u}\left(t\right), \bm{\theta}\right),
\end{equation}

\noindent where $\bm{\theta}$ is the collection of all trainable weights in the KAN. Here, the input and output dimensions of the $\text{KAN}$ are fixed as the dimension of the state variables $\mathbf{u}$, and it is used as a gradient getter. Equation~\ref{eq:KAN_gradient} can be solved using a traditional ODE solver from an initial time $t_{0}$ to a desired time $t$ such that

\begin{equation} \label{eq:KAN_ODE}
\mathbf{u}(t) = \mathbf{u}_{0} + \int_{t_{0}}^{t} {\text{KAN} \left(\mathbf{u}\left(\tau\right), \bm{\theta}\right)d\tau },
\end{equation}

\noindent where $\mathbf{u}_{0}$ is the initial condition at $t_0$.

\begin{figure}[tb]
    \centering
	\includegraphics[width=\linewidth]{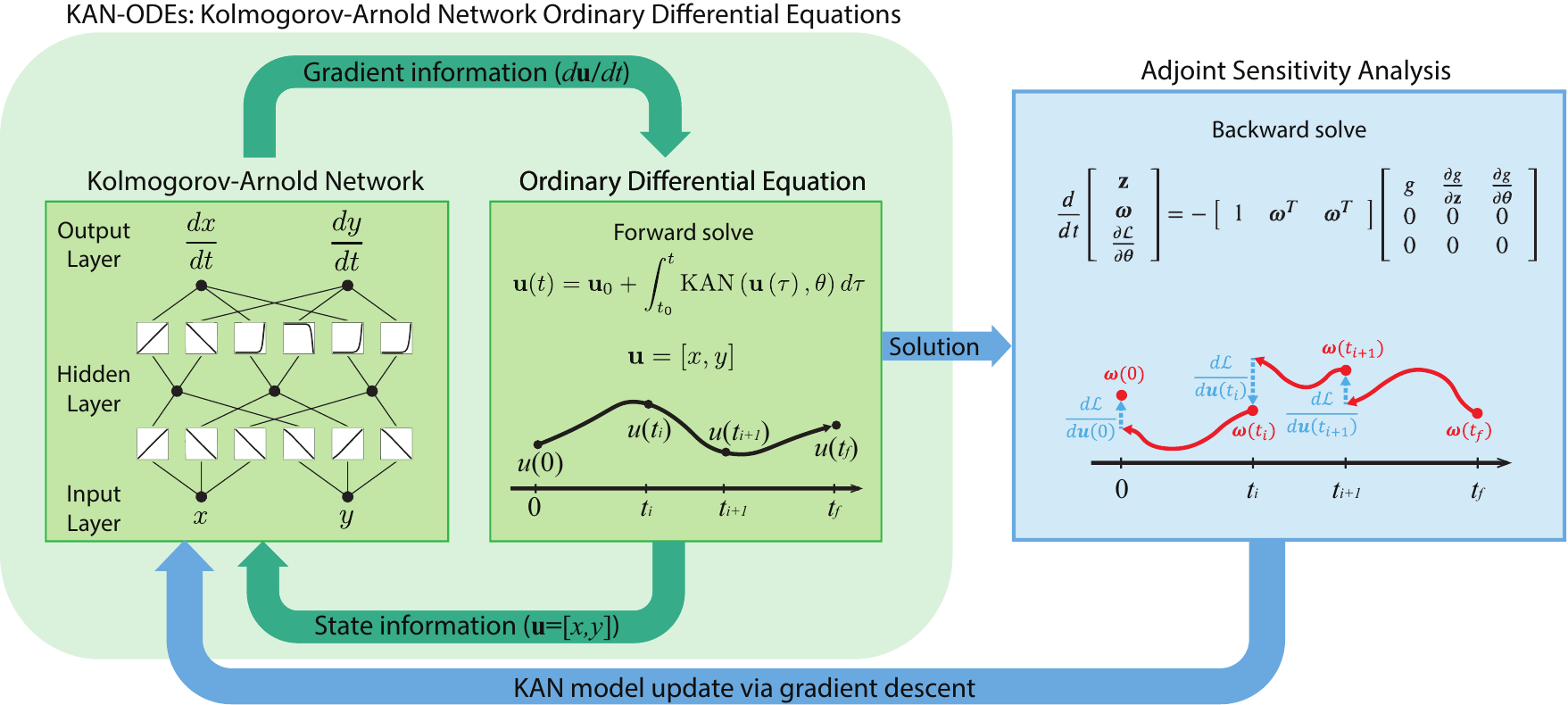}
	\caption{Schematic depicting the overall training cycle of a KAN-ODE. The loop in green leverages a KAN as a temporal gradient getter for the state vector to solve the ODE forward. Once a solution is generated, the blue loop computes the gradient of the loss function via the adjoint method to update the KAN activation functions.}
	\label{fig:schematic}
\end{figure}

To train the \text{KAN} model, we define the loss function as the mean squared error

\begin{equation}\label{eq:loss}
    \mathcal{L}\left(\bm{\theta} \right) = \text{MSE}\left(\mathbf{u}^{\text{KAN}}\left(t,  \bm{\theta}\right), \mathbf{u}^{\text{obs}}\left(t\right) \right) = \frac{1}{N}\sum_{i=1}^N {\lvert \lvert\mathbf{u}^{\text{KAN}}\left( t_{i}, \bm{\theta}\right)- \mathbf{u}^{\text{obs}}\left(t_{i}\right)\rvert \rvert ^{2}},
\end{equation}

\noindent where $\mathbf{u}^{\text{KAN}}$ is the prediction, $\mathbf{u}^{\text{obs}}$ is the observation, and $N$ is the total number of time steps. To optimize $\mathcal{L}$, we require gradients with respect to $\bm{\theta}$. Generally, two options are available: the \textit{forward} sensitivity method \cite{kim_stiff_2021,kim2023inference} and the \textit{adjoint} sensitivity method \cite{chen_neural_2019,rackauckas2020universal}. In this paper, the \textit{adjoint} sensitivity method is adopted to efficiently handle a potentially large Kolmogorov-Arnold network \cite{rackauckas2020universal,kim_stiff_2021, chen_neural_2019} thanks to its more advantageous scaling with model size. By introducing an adjoint state variable $\bm{\omega}$, the augmented system equation can be derived, which allows for gradient backpropagation through the ODE integrators.

\begin{equation}\label{eq:adjoint}
\frac{d}{dt}\left[\begin{array}{c}
\mathbf{z}\\
\bm{\omega}\\
\frac{\partial \mathcal{L}}{\partial \bm{\theta}}
\end{array}\right]=-\left[\begin{array}{ccc}
1 & \bm{\omega}^{T} & \bm{\omega}^{T}\end{array}\right]\left[\begin{array}{ccc}
\mathbf{g} & \frac{\partial \mathbf{g}}{\partial \mathbf{z}} & \frac{\partial \mathbf{g}}{\partial \bm{\theta}}\\
0 & 0 & 0\\
0 & 0 & 0
\end{array}\right],
\end{equation}

\noindent where $\mathbf{z}(t)=\mathbf{u}(-t)$, $\mathbf{g}=\text{KAN}$, and $T$ denotes the transpose. \hl{Detailed derivations of the adjoint equations (Eq.~\mbox{\ref{eq:adjoint}}) can be found in \mbox{\cite{maly1996numerical,cao2002adjoint,cao2003adjoint,chen_neural_2019}}.} After Eq.~\ref{eq:adjoint} is computed, $\bm{\theta}$ is updated using a gradient descent method. A schematic depicting the overall training cycle is provided in Fig. \ref{fig:schematic}.

This study implements KAN-ODEs in the Julia scientific machine learning ecosystem including packages such as DifferentialEquations.jl \cite{rackauckas_differentialequationsjl_2017}, Lux.jl \cite{pal2023lux}, KomolgorovArnold.jl \cite{puri_kolmogorovarnoldjl_2024}, and Zygote.jl. Unless otherwise specified, we employed the ODE integrator of \verb|Tsit5| (Tsitouras 5/4 Runge-Kutta method \cite{tsitouras_rungekutta_2011}) and the optimizer of \verb|ADAM| \cite{kingma_adam_2017}. In addition, we constructed \text{KANs} with RBF basis functions and Swish residual activation functions.

\section{\label{sec:results_overview}Experiments}
We present five inference examples in this work to demonstrate the broad capability of KAN-ODEs, grouped into three subsections. We first validated our methodology and compared its performance against the standard MLP-based Neural ODE in Sec.~\ref{sec:results_LV}, where the dynamics of the Lotka-Volterra predator-prey model were inferred. There we also comprehensively explored the neural scaling benefits of KAN-ODEs, as well as the efficient representations that can be inferred via sparsification and pruning. Then, we tackled a one-dimensional PDE in Sec.~\ref{sec:results_wave}, using a KAN-ODE to infer a symbolic source term in a wave propagation simulation. This example demonstrates the flexibility of KAN-ODEs to be used as submodels in higher complexity simulations, as well as their capability to directly extract symbolic models from data. Finally, we demonstrated the scalability and inference power of KAN-ODEs as standalone models in Sec.~\ref{sec:results_burg} by inferring complete data-driven solutions to PDEs using limited samples. Examples there include shock wave formation governed by the Burgers' equation and quantum wave function evolution governed by the Schr\"{o}dinger equation, with an additional phase separation case using the Allen-Cahn equation provided in Appendix C.

\subsection{\label{sec:results_LV}KAN-ODEs vs Neural ODEs: Extensive Comparison via Lotka-Volterra Equations}
In this section, we targeted the Lotka-Volterra predator-prey model shown in Eq.~\ref{eq:LV}. We replicated the simulation parameters used in the Neural ODE study of \cite{dandekar_bayesian_2022}. Namely, we used $\alpha=1.5$, $\beta=1$, $\gamma=1$, and $\delta=3$ with an initial condition of $\mathbf{u}_{0}=[x_0, y_0]=[1, 1]$ and a time span of $t \in [0, 14]$ s. \hl{The temporal grid used for data generation, training, and loss evaluation had a spacing of 0.1 s. The} {\verb|Tsit5|} \hl{ODE integrator was used to solve the governing equations for the true ODE trajectory (i.e. the "ground truth" data used to train and validate the KAN-ODE model). The first 3.5 s of this ground truth trajectory was used to train the KAN-ODE (Fig.~{\ref{fig:LV}}(A)) via Eqs. {\ref{eq:loss}} and {\ref{eq:adjoint}}, while the remainder of the time history (3.5 s to 14 s) was withheld to validate the KAN-ODE's ability to forecast the states at unseen times. While the training data is reported at 0.1 s increments, the KAN-ODE's use of the} {\verb|Tsit5|} \hl{adaptive solver requires much finer fidelity in its gradient accuracy.} We tested the KAN-ODEs with various architectures shown in Table~\ref{table_LV_body}. The inputs to each KAN layer in this example were normalized to be on the $[-1,1]$ range required by the RBF networks via the hyperbolic tangent function, as in \cite{puri_kolmogorovarnoldjl_2024}.

\begin{equation}
\begin{aligned}\label{eq:LV}
\frac{dx}{dt} &= \alpha x - \beta xy,\\
\frac{dy}{dt} &= \gamma x y - \delta y.
\end{aligned}
\end{equation}

Loss profiles from training a 240-parameter KAN-ODE of shape [2, 10, 5], [10, 2, 5] as a representative case are shown in Fig. \ref{fig:LV}(B1). Strong convergence is observed at or even before $10^{4}$ epochs into training, while convergence into the $10^{-7}$ range of MSE loss occurs toward $10^{5}$ epochs. The converged KAN-ODE prediction is shown in Fig.~\ref{fig:LV}(A), where the agreement in the testing window ($t\in[3.5,14]$) is nearly indistinguishable from the agreement in the training window ($t\in[0,3.5]$). We remark again that the trained KAN-ODE is reconstructing both the seen training data \hl{(i.e. the ground truth ODE solution from 0 s to 3.5 s) and unseen testing data (3.5 s to 14 s)} given only the initial condition ($\mathbf{u}_{0}=[1, 1]$), and can deliver output values on any temporal grid, thanks to its coupling to an ODE solver. 

To contextualize these results, benchmark tests were performed to compare KAN-ODEs with standard, MLP-based Neural ODEs. We draw comparisons in terms of accuracy, convergence speeds, and neural scaling in Sec.~\ref{sec:LV_scaling}; and interpretability and generalization in Sec. \ref{sec:LV_generalize}. In Sec.~\ref{sec:LV_generalize}, we additionally discuss the impacts of sparsified and symbolic KAN-ODEs.

\subsubsection{Benchmark Tests and Neural Scaling Behavior} \label{sec:LV_scaling}

\begin{figure*}[tb]
    \centering
	\includegraphics[width=0.9\linewidth]{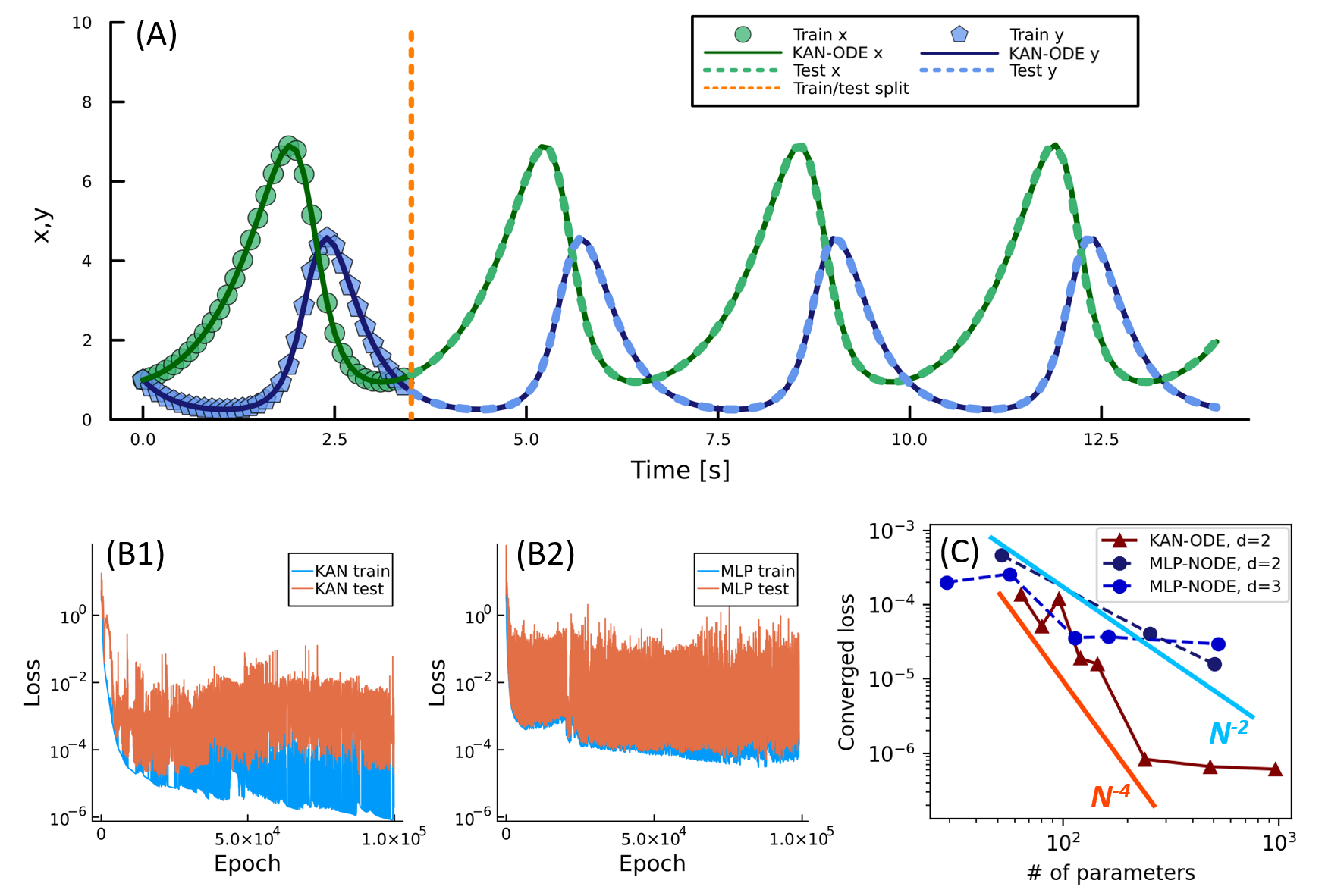}
	\caption{Comparison between KAN-ODE and Neural ODE for Lotka-Volterra predator-prey model. (A) Synthetic data (training and testing) and KAN-ODE reconstruction. (B) Loss profile during training for (B1) KAN-ODE and (B2) MLP-based Neural ODE of comparable sizes (240 and 252, respectively). (C) Comparison of converged KAN-ODE and Neural ODE error using different model sizes, and two MLP depths ($d=2$ and $d=3$). Neural scaling rates of $N^{-2}$ and $N^{-4}$ plotted for comparison, as per the theory in \cite{liu_kan_2024}.}
	\label{fig:LV}
\end{figure*}

The MLP size and training procedure were taken from the Neural ODE work of \cite{dandekar_bayesian_2022} to avoid biasing the results in the KAN-ODE's favor. Namely, an MLP with a hidden layer comprising 50 nodes and the hyperbolic tangent activation function replaced the KAN in Eq.~\ref{eq:KAN_gradient}. This MLP contained 252 trainable parameters (refer to the bold architecture in Table~\ref{table_LV_body}), which is comparable (and in fact slightly larger) than the 240 used in the reported KAN-ODE, facilitating fair comparison. Other relevant training details such as the train/test time windows, temporal grid, Lotka-Volterra equation parameters, and ODE solver remained identical to the KAN-ODE example, which itself was modeled as directly as possible after the Neural ODE efforts of \cite{dandekar_bayesian_2022}.

\begin{table*}[tb]
\caption{Lotka-Volterra tests with Neural ODEs and KAN-ODEs of varying size. Bold entries are studied further in Fig.~\ref{fig:LV}.}
\begin{ruledtabular}
    
\begin{tabular}{ccccccc} 
  &  Depth  &  Layer width  &  Grid size &  Activation Function  &  No. Params  &  Train loss  \\ [0.5ex] 
 \hline
 
Neural ODE (MLP) & 2  & 10&N/a & tanh & 52&$4.7 \times 10^{-4}$  \\
&  \textbf{2} &  \textbf{50}& \textbf{N/a}  & \textbf{tanh}&  \textbf{252} &$\mathbf{4.1 \times 10^{-5}}$  \\
& 2  & 100&N/a & tanh & 502 &$1.6 \times 10^{-5}$  \\

& 3  & 3&N/a& tanh  & 29&$2.0 \times 10^{-4}$  \\
& 3  & 5&N/a& tanh  & 57&$2.6 \times 10^{-4} $ \\
& 3  & 8&N/a & tanh & 114&$4.6 \times 10^{-5}$  \\
& 3  & 10&N/a& tanh  & 162&$3.7 \times 10^{-5} $ \\
& 3  & 20&N/a  & tanh& 522&$3.0 \times 10^{-5} $ \\
 \midrule

 KAN-ODE & 2  & 4&3 & \textit{learned}& 64&$1.4 \times 10^{-4}$  \\
 & 2& 4&4  & \textit{learned}& 80 &$5.2 \times 10^{-5}$ \\
 & 2  & 4&5 & \textit{learned}& 96&$1.2 \times 10^{-4} $ \\
 & 2  & 6&4 & \textit{learned}& 120&$1.9 \times 10^{-5}$  \\
 & 2  & 6&5& \textit{learned}& 144 &$1.6 \times 10^{-5} $ \\
 &  \textbf{2} &  \textbf{10}& \textbf{5} & \textbf{\textit{learned}} & \textbf{240}&$ \mathbf{8.3 \times 10^{-7}}$  \\
 & 2  & 20&5 & \textit{learned}& 480&$6.6 \times 10^{-7} $ \\
 & 2  & 40&5 & \textit{learned}& 960&$6.1 \times 10^{-7} $ \\

\end{tabular}
\end{ruledtabular}

\label{table_LV_body}
\end{table*}

In this 1-to-1 comparison using a data generation and training regime borrowed from a well-cited Neural ODE work, our proposed KAN-ODE beats the performance of the Neural ODE in all key metrics. Inspection of the loss profiles of Figs. \ref{fig:LV}(B1-B2) shows that the KAN-ODE converged to a substantially smaller training loss of $8.3\times10^{-7}$ over $10^{5}$ epochs compared to the similarly-sized Neural ODE, which reached only $3\times10^{-5}$ in the same number of epochs. \hl{Both loss profiles exhibit significant oscillatory behavior while still successfully training, which we theorize may be due to the data-lean, large-batch regime studied here (i.e. a single dataset, all time steps of which are optimized in one batch). These oscillations, which do not appear to affect the final result, are smaller in the KAN-ODE training cycle than in the Neural ODE training cycle.} The KAN-ODE qualitatively appears to overfit more than the Neural ODE model, especially in the later epochs. Quantitatively, however, it still beats the Neural ODE in accuracy to the unseen testing data, \hl{indicating an objective improvement in performance when switching from the Neural ODE to the KAN-ODE}. Even in the last 10\% of the training cycle (epochs $90,000-100,000$), where the KAN-ODE \hl{experiences a notable difference between training and testing losses,} it still achieved an average testing loss of $6.8\times10^{-3}$, with a minimum value in that range of $1.9\times10^{-5}$. The Neural ODE in this same range averaged a much larger $1.4\times10^{-2}$, with a minimum value of $5.4\times10^{-5}$. If training for the KAN-ODE was stopped early at $2\times10^4$ epochs when the testing error just begins to creep upward, these statistics would tend even more strongly in its favor. At this early stopping point, \hl{little to no indication of overfitting is observed when comparing its own training and testing losses, and all loss metrics still outperform those of the Neural ODE (even when the latter is trained to the full $1\times10^5$ epochs). The KAN-ODE solution of Fig. {\ref{fig:LV}(A)} can additionally be generated at arbitrary time steps (e.g. 0.01 s rather than 0.1 s) with no detectable change to the testing results, indicating no overfitting to the specific training timesteps used here.}

In terms of computational cost, we found that the KAN-ODE iterated relatively slowly but actually converged 3-4x faster than the Neural ODE. On a single CPU core, the 240-parameter KAN-ODE took around 19 minutes per $10^{4}$ epochs, compared to the 252-parameter Neural ODE which took closer to 7 minutes per $10^{4}$ epochs. However, the KAN-ODE reached $2.6\times10^{-5}$ training loss in just $10^{4}$ epochs, already beating the best-performing, $10^{5}$ epoch Neural ODE result of $3.0\times10^{-5}$ training loss. Thus, while the KAN-ODE iterated 2.5 to 3x slower than the Neural ODE, it was able to reach the same predictive performance with 10x fewer epochs, resulting in an overall effective 3-4x speedup over the Neural ODE.

\begin{figure*}[tb]
    \centering
	\includegraphics[width=0.95\linewidth]{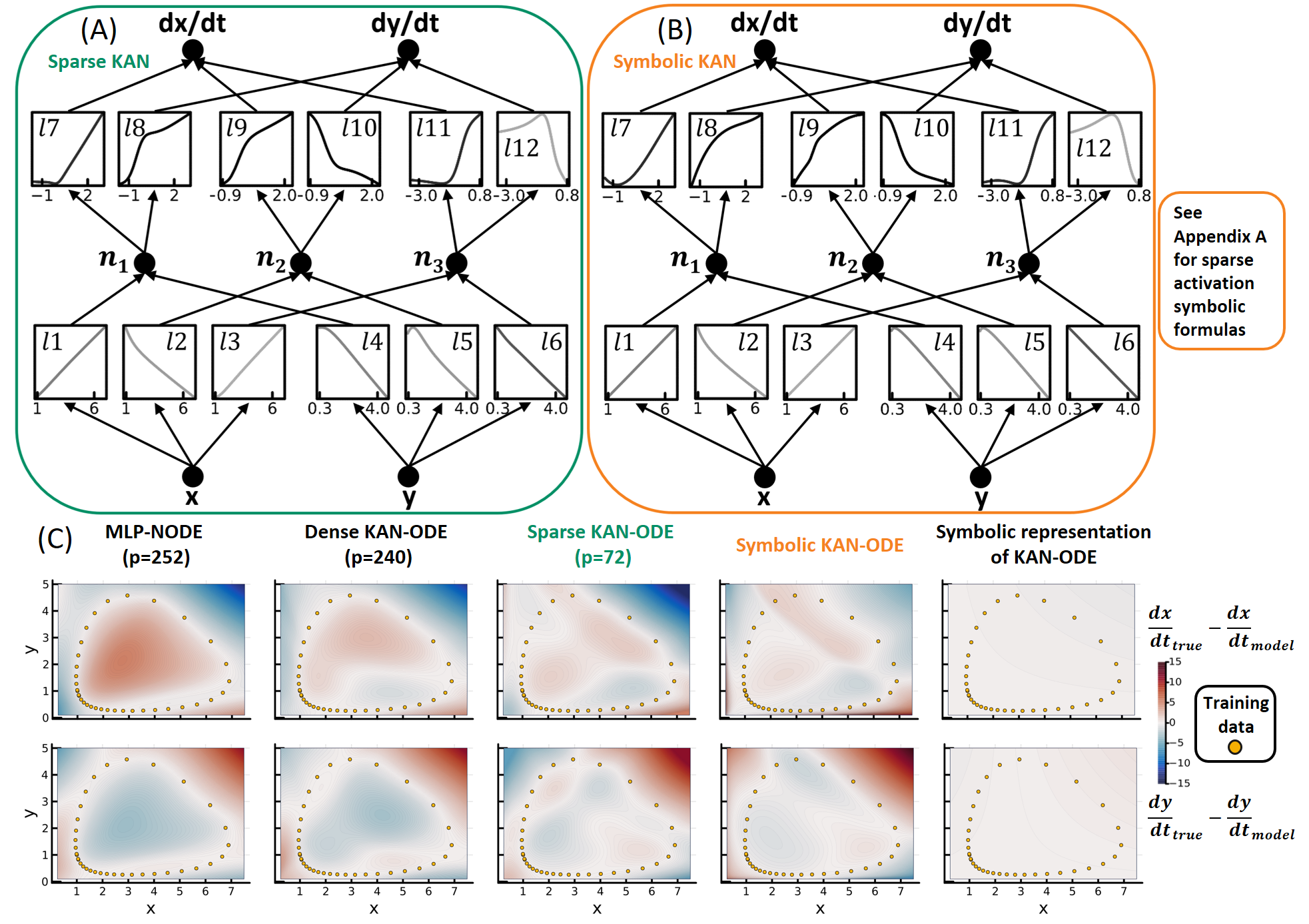}
	\caption{Sparsification, pruning, symbolic regression, and generalization results for Lotka-Volterra dynamics. (A) Sparse KAN with 72 parameters ([2, 3, 5], [3, 2, 5]), pruned from an initial 240 parameters ([2, 10, 5], [10, 2, 5]). (B) Symbolic KAN derived from (A), where each of the twelve activations is replaced by a univariate symbolic expression. (C) Generalization error when extrapolating outside of the yellow $(x, y)$ points explored in the training data. Each column represents a different gradient getter and has one row each for $dx/dt$ and $dy/dt$. Generalization is seen to improve when moving from MLP gradients to increasingly sparse KANs, and finally to symbolic representations fitted to KANs.}
	\label{fig:LV2}
\end{figure*}

 Up to this point, comparisons have been drawn exclusively between a single Neural ODE architecture and a single KAN-ODE architecture. Here, we discuss the effects of varied network architectures on the performance of KAN-ODEs and Neural ODEs, the results of which demonstrate a key scaling improvement of KAN-ODEs. Liu et al. \cite{liu_kan_2024} proposed that, thanks to the third-order splines used in the Kolmogorov-Arnold representations, KANs benefit from fourth-order neural scaling, or a quartic decrease in loss with respect to the KAN size (parameterized here as the number of trainable parameters). MLPs, on the other hand, often saturate quickly and can struggle to reach even first-order convergence \cite{liu_kan_2024, michaud_precision_2023}. We investigated this phenomenon here by varying the layer width and grid size of KAN-ODEs, and by varying the layer width and depth of MLP-based Neural ODEs. Figure \ref{fig:LV}(C) shows that, even with the modified RBF basis function, we appear to capture the $N^{-4}$ B-spline KAN scaling rate with KAN-ODEs, while the two depths of MLP appear to be converging at at rate slower than $N^{-2}$. With just the 240 KAN parameters studied in Figs. \ref{fig:LV}(A) and \ref{fig:LV}(B1) we were able to run the fourth-order KAN-ODE convergence to saturation (larger KANs see no substantial additional performance gain), while MLP-based Neural ODEs would theoretically require upwards of thousands or even tens of thousands of parameters to reach similar performance, if they did not plateau earlier. The specific model sizes used in this convergence test are shown in Table \ref{table_LV_body}, where the models used for Figs. \ref{fig:LV}(A) and \ref{fig:LV}(B1-B2) are in bold. To summarize, we found here that KAN-ODEs beat Neural ODEs in terms of data-fitting performance and computational efficiency in a systematic study of a broad range of network architectures and sizes.

\subsubsection{Interpretation and Generalization of KAN-ODEs with Varying Sizes} \label{sec:LV_generalize}

In this section we implemented sparsification, pruning, and symbolic regression to demonstrate a framework for developing interpretable KAN-ODE solvers. We additionally heuristically explored the generalizability of the various MLP and KAN-based models, where we found that KAN-ODEs have strong generalization potential even with limited training data. In addition to the 252-parameter Neural ODE and 240-parameter KAN-ODE (trained in Sec. \ref{sec:LV_scaling}) as reference cases, three models were studied: a sparse KAN-ODE trained via parameter regularization and pruning, a symbolic KAN-ODE developed via symbolic regression on each univariate activation of the sparse KAN-ODE, and a symbolic representation for the entire model via symbolic regression on the global input-output pairs of the sparse KAN-ODE.

Sparsification, as proposed in \cite{liu_kan_2024} and discussed in \cite{blealtan_efficient-kan_2024}, constitutes an additional term in the loss function to penalize non-zero trainable parameters, driving a given model toward sparser representations with fewer large activations. Here we use the simple L1 norm,

\begin{equation} \label{eq:sparse_loss}
    \mathcal{L}\left(\bm{\theta} \right) = \text{MSE}\left(\mathbf{u}^{\text{KAN}}\left(t,  \bm{\theta}\right), \mathbf{u}^{\text{obs}}\left(t\right) \right) + \gamma_{sp} |\bm{\theta}|_1.
\end{equation} 
This is identical to Eq. \ref{eq:loss}, with the inclusion of the L1 penalty term with $\gamma_{sp}$ as a hyperparameter, which drives unnecessary activations toward zero. The logical next step is to reduce the size and complexity of the now-sparse KAN-ODE by eliminating any trivial nodes via pruning. To do so, we compare the training inputs and outputs of each node against a pruning parameter, $\gamma_{pr}$. Any node where the magnitude of at least one input or output exceeds $\gamma_{pr}$ is kept in the KAN-ODE, while those with all inputs and outputs smaller than $\gamma_{pr}$ are pruned. We began with a [2, 10, 5], [10, 2, 5] structure for a KAN (referred to henceforth as the dense KAN), based on observation from Fig. \ref{fig:LV}(C) of this architecture's optimal performance. To sparsify the KAN-ODE, we included the regularization term of Eq. \ref{eq:sparse_loss} with $\gamma_{sp}=5 \cdot 10^{-4}$ in the training process. To prune, we trained this KAN-ODE to convergence, then eliminated nodes that did not satisfy $\gamma_{pr}=10^{-2}$ before training the remaining KAN-ODE to convergence once again. Figure \ref{fig:LV2}(A) shows the sparse KAN-ODE with a final pruned size of [2, 3, 5], [3, 2, 5] trained using this procedure. Note that the darkness of the activation function curves is determined by the ratio of the outputs to the inputs (larger outputs have thicker lines).

Next, the symbolic KAN-ODE took the sparse KAN and fit symbolic expressions to each of its twelve univariate activation functions using the SymbolicRegression.jl package \cite{cranmerInterpretableMachineLearning2023} with the four fundamental binary operators $[+,-,\times,/]$. Detailed results for each of these regressions is available in the Appendix A. Here, we plot the twelve symbolic activations for visualization in Fig. \ref{fig:LV2}(B). To summarize, we first sparsified a 240-parameter, 10-node KAN into a 72-parameter, 3-node KAN. Then, we replaced these 72 parameters with just twelve univariate activation functions, a handful of which in the first layer are simple linear functions of the form $y=mx+b$. In terms of interpretability, we moved from a black box MLP-based Neural ODE to a visualizable yet bulky KAN, then to a visualizable and compact KAN, and finally to a straightforward collection of symbolic relationships that capture the Lotka-Volterra dynamics well. None of these steps required knowledge of the system outside of the provided training data.

Finally, we ran the entire Sparse KAN through SymbolicRegression.jl (rather than each activation separately) and found the following relationships which comprise the KAN representation,

\begin{equation}
\begin{aligned}\label{eq:symbolic_LV}
\frac{dx}{dt} &= 1.495 x - 0.986 xy,\\
\frac{dy}{dt} &= 0.970 x y - 2.929 y,
\end{aligned}
\end{equation}

\noindent which correspond nearly identically to our model parameters of 1.5, 1, 1, and 3. This near-perfect result is effectively the pinnacle of interpretability and accuracy, though we believe is not a realistic option for most systems of larger-scale merit. Complete symbolic regression on an entire learned model or dataset is often not feasible, especially for complex models with higher dimensional inputs and outputs than are studied in this relatively straightforward ODE system \cite{liu_kan_2024}. Even when numerically manageable, clean convergence to compact functional forms is not guaranteed. On the other hand, a symbolic KAN like that shown in Fig.~\ref{fig:LV2}(B) requires symbolic regression only on a series of straightforward univariate activations and retains entirely interpretable intermediate steps. We present this global symbolic representation result as an interesting aside and a benchmark for comparison, but propose that sparse KAN-ODEs and symbolic KAN-ODEs are the most likely to have broader applicability in more realistic systems depending on the desired degree of interpretability and accuracy.

We move next to a discussion of the generalizability of KAN-ODEs. To conform to prior studies and faciliate fair comparison against MLP-based Neural ODEs, all models used previously in this Lotka-Volterra section were trained on data generated from a single initial condition, as was done in the Neural ODE study of \cite{dandekar_bayesian_2022}. There exist many other potential $(x, y)$ combinations, however, and here we heuristically tested these trained models on their ability to generalize to completely unseen conditions.

The gradient error landscapes of the five models studied here (MLP-NODE, Dense KAN-ODE, Sparse KAN-ODE, Symbolic KAN-ODE, and symbolic representation of KAN-ODE) are shown from left to right, respectively, in Fig. \ref{fig:LV2}(C), where the top row is the $dx/dt$ error and the bottom row is the $dy/dt$ error. The MLP-NODE results in the first column show near-zero error in the immediate vicinity of the training data, as expected from the performance metrics of Fig. \ref{fig:LV}. However, they also show significant gradient error in almost all ($x, y$) ranges not directly on or immediately near the training data points, revealing the expected weak generalizability of the standard Neural ODE due to the fact that training data is limited to the marked points only.

The gradient error landscape of the dense KAN-ODE beats the MLP in generalizability, with much smaller gradient error values throughout the domain and especially inside of the yellow training points where not only is the overall error substantially reduced, but we begin to additionally observe the formation of thin bands of near-perfect reconstruction away from the data points. This result is rather interesting, as it was originally proposed in \cite{liu_kan_2024} that KANs beat MLPs in generalization thanks to their smaller number of parameters, but we observe here a substantial improvement with similarly sized models. In any case, these results suggest the potential of KAN-ODEs in the development of powerful models from limited data. \hl{In a brief aside, we also remark on the current result's support of our previous argument (in Sec. {\ref{sec:LV_scaling}}) that the fully-trained dense KAN-ODE does not overfit to the training data points, given its near-perfect interpolation between these points along the training profile. This result is unsurprising given the adaptive time stepping of the} {\verb|Tsit5|} \hl{solver, but is nonetheless a strong result, especially when combined with the additional visible improvement in generalization outside of the training profile. If overfitting becomes substantial in larger-scale KAN-ODE applications, there are a handful of general neural network techniques and Neural-ODE adapted techniques that can be adapted to address this \mbox{\cite{liu_neural_2019, ying_overview_2019, srivastava_dropout_2014}} including early stopping and network reduction (as proposed here) as well as data expansion, noise injection, and dropout.}

The gradient error landscape of the sparse KAN-ODE shows improvements over the dense KAN-ODE landscape. Inside the yellow data points, we see further formation of near-perfect gradient reconstruction bands, surrounded by relatively low error regions. Results toward the edges of the landscape are mixed: the $dx/dt$ reconstruction captures the top left of the ($x, y$) domain with excellent accuracy but suffers from a large error in the upper right portion of the domain, while the $dy/dt$ reconstruction performs well toward low $y$ values, but suffers at high $y$ values. On the whole, we remark an overall improvement in generalization with a nearly 4x reduction in KAN size, indicating the utility of sparse KAN-ODEs not just for compact and interpretable models, but also for improved generalization and extrapolation capabilities thanks to their smaller size. This result corroborates similar observations from \cite{liu_kan_2024} of the increase in generalizability with decreased KAN parameter counts.

The Symbolic KAN-ODE sees remarkable improvement in the $dx/dt$ generalization, with the low-error area being pushed out significantly toward high $x$ and $y$ values, further growth of the white bands of near-zero error throughout the domain, and a general lightening of the plot indicating better extrapolation and generalization. The $dy/dt$ reconstruction sees modest improvement, especially inside the data points. We propose that this generalization improvement occurs due again to the more compact forms present in the symbolic KAN-ODE when compared to even the sparse KAN-ODE, and the extrapolation benefits that come with smaller system representations \cite{liu_kan_2024}. In the significantly out-of-training regions of $dx/dt$ where the symbolic KAN-ODE succeeds and the gridded basis functions of the dense KAN-ODE and Sparse KAN-ODE fail, we suggest that the input samples may be exiting the trained region of the RBF basis functions yet remaining within the useful range of the expressions used in the symbolic KAN-ODE.

The symbolic representation of the entire KAN obviously generalizes with near-perfect accuracy thanks to its discovery of the exact underlying model form (with near-perfect parameters). This result is plotted in the final column of Fig. \ref{fig:LV2}(C) for reference, but we again emphasize that the other three KAN models are more reasonable choices for realistic applications: Dense KAN-ODEs succeeded with remarkably low training and testing error, sparse KAN-ODEs retained much of this low error while appearing to improve generalizability thanks to their smaller size, and Symbolic KAN-ODEs were substantially more interpretable and human-readable while also appearing to improve generalizability thanks to their smaller size and replacement of limited-scope basis functions with standard symbolic functions.

In this Lotka-Volterra example, we systematically and comprehensively explored the use of KAN-ODEs in a straightforward ODE system and benchmarked their performance against comparable MLP-based Neural ODEs in terms of accuracy, speed, model size, convergence rate, interpretability, and generalization. We found that KAN-ODEs are an extremely efficient method to represent the predator-prey dynamics of this case, and beat Neural ODEs in all relevant quantitative metrics. We additionally explored the interpretability of KAN-ODEs via symbolic regression of each activation function, and then heuristically evaluated a series of MLP and KAN-based ODE approaches in terms of generalizability to unknown data, where we again found KAN-ODEs superior. In the next sections, we present larger-scale examples to showcase the flexibility and inference capabilities of KAN-ODEs.

\subsection{\label{sec:results_wave}Modeling Hidden Physics in PDEs: Fisher-KPP PDE}

In this section, we introduce the capability of KAN-ODEs to learn the hidden physics in PDEs. Key features demonstrated here are the flexibility of KAN-ODEs to be paired with higher-complexity solvers on arbitrary temporal grids, and the capability of KAN-ODEs to directly extract hidden symbolic functional relationships from training data. We demonstrate this using the Fisher-KPP equation representing a reaction-diffusion system, 

\begin{equation}\label{Eqn:Fisher-KPP}
\frac{\partial u}{\partial t} = D \frac{\partial^2 u}{\partial x^2} + ru(1-u), 
\end{equation}

\noindent where $D$ is the diffusion coefficient and $r$ is  the local growth rate. We assume that the reaction term of $ru(1-u)$ is unknown and to be modeled with a KAN. 

Training data was synthesized by solving Eq.~\ref{Eqn:Fisher-KPP} in the domain of $x\in[0,1]$, $t\in[0,5]$, given the model parameters $D=0.2$ and $r=1.0$. The initial condition and boundary conditions were

\begin{align}
    u(x,0)&=\frac{1}{2}\left[\tanh\left(\frac{x - 0.4}{0.02}\right) - \tanh\left(\frac{x - 0.6}{0.02}\right)\right],\\
    u(0,t)&=u(1,t),\\
    \frac{\partial u}{\partial x}\left(0,t\right)&=\frac{\partial u}{\partial x}\left(1,t\right) \notag.
\end{align}

\begin{figure*}[tb]
    \centering
	\includegraphics[width=\linewidth]{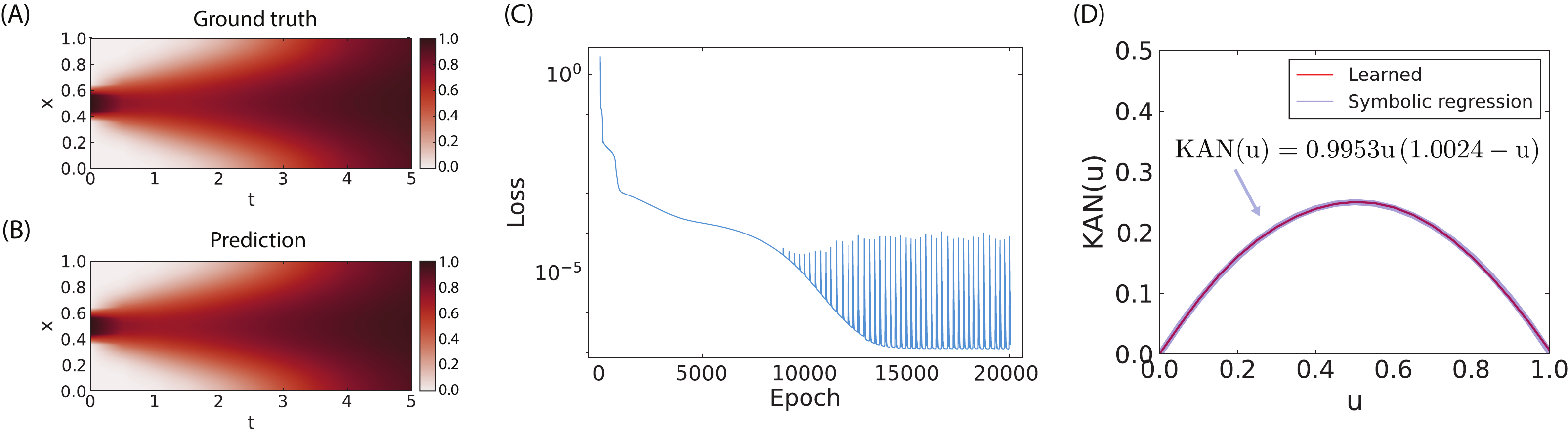}
	\caption{\textit{Fisher-KPP equation}: (A) Solution field $u(x,t)$ of the ground truth. (B) Solution field $u(x,t)$ of the prediction with a learned KAN-ODE model. (C) Loss function. (D) Learned hidden physics of the reaction source term in the Fisher-KPP equation and its symbolic form.}
	\label{fig:wave}
\end{figure*}

\noindent Note that this problem setup was adopted from \cite{dandekar_bayesian_2022}. The spatial grid was discretized with $\Delta x=0.04$, and the equation was solved with a time step of $\Delta t=0.5$ \hl{using the ODE integrator of} \verb|Tsit5| \hl{(Tsitouras 5/4 Runge-Kutta method \mbox{\cite{tsitouras_rungekutta_2011}})}. The obtained solution field $u(x,t)$ is shown in Fig.~\ref{fig:wave}(A) and used as training data.

To model the unknown reaction source term in Eq.~\ref{Eqn:Fisher-KPP}, we formulated the KAN-ODE such that

\begin{equation}\label{Eqn:Fisher-KPP-KAN}
\frac{\partial u}{\partial t} = D \frac{\partial^2 u}{\partial x^2} + \text{KAN}\left(u, \bm{\theta}\right). 
\end{equation}

\noindent To utilize backpropagation with the \textit{adjoint} sensitivity method, Eq.~\ref{Eqn:Fisher-KPP-KAN} was discretized using a central difference scheme and converted to an ODE system using the Method of Lines. The KAN was constructed with a single layer comprising a single node, [1,1,10].

The prediction with a learned model is illustrated in Fig.~\ref{fig:wave}(B), showing accurate reconstruction of the solution field $u(x,t)$ after as few as 5,000 updates (Fig.~\ref{fig:wave}(C)) and demonstrating that a single learned activation function comprised of ten gridded basis functions ([1,1,10]) is sufficient to represent the reaction source term (see Appendix B for a detailed discussion of accuracy with varying KAN-ODE architectures). This resultant activation function is shown in Fig.~\ref{fig:wave}(D). We further propose expressing this single activation function symbolically, as was introduced in \cite{liu_kan_2024} as a feature of KAN layers. Symbolic regression on this single KAN-ODE activation using SymbolicRegression.jl \cite{cranmerInterpretableMachineLearning2023} given the candidate symbolic expressions of $[+,-,\times,/,\sin, \cos, \exp]$ returned the following expression,

\begin{equation}
    \text{KAN}\left(u\right) = 0.995311u(1.002448-u).
\end{equation}

\noindent Given its single-layer, single-node construction, this symbolic expression is the KAN-ODE's approximation of the true reaction source term in Eq.~\ref{Eqn:Fisher-KPP}. Knowing $r=1.0$, we see that the KAN-ODE's derived expression is remarkably close to the original formulation. Thus, the KAN-ODE was not only capable of learning the dynamics of physics hidden within a PDE submodel from measurable quantities but also of returning an accurate and human-interpretable symbolic function that mimics the true governing law with excellent accuracy.

\subsection{\label{sec:results_burg}Data-Driven Solutions of PDEs}

\subsubsection{Burgers' Equation}

Our third example explores the use of a KAN-ODE to infer the unknown hidden states $u\left(x,t\right)$ of a PDE system, effectively representing all of its spatiotemporal dynamics. Here we considered the Burgers' convection-diffusion equation as an example.

\begin{equation}\label{eq:burger}
   \frac{du}{dt} + u  \frac{du}{dx} = \frac{0.01}{\pi}\frac{d^{2}u}{dx^{2}}.
\end{equation}

\noindent The computational domain was defined as $x \in [-1.0, 1.0]$ and $t\in [0.0, 1.0]$.

In order to prepare training data, Eq.~\ref{eq:burger} was discretized using a central difference scheme with $\Delta x = 0.05$ such that $\mathbf{u}\left(t\right)=\left[u\left(x_{0},t\right), u\left(x_{1},t\right), ..., u\left(x_{N},t\right)\right]$. Then, the discretized PDE was solved by an ODE integrator given the initial and boundary conditions such that

\begin{align}
	u\left(x,0\right)&=-\sin (\pi x),\label{eq:burger_ic}\\
	u\left(-1,t\right)&=0,\label{eq:burger_bc}\\
	u\left(1,t\right)&=0. \notag
\end{align}

\begin{figure*}[tb]
	\includegraphics[width=\linewidth]{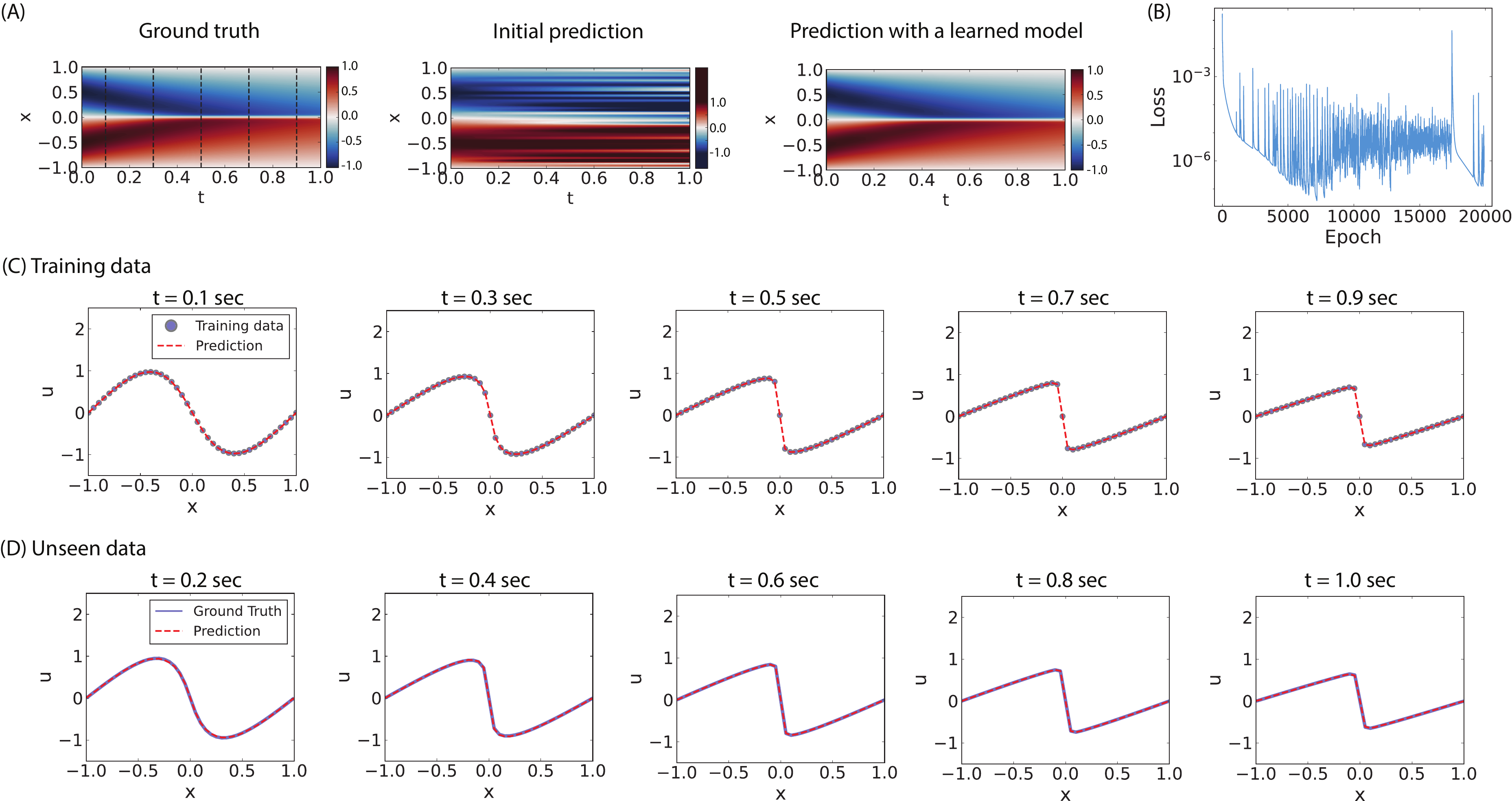}
	\caption{\textit{Burgers' equation}: (A) Solution fields $u(x,t)$ for the ground truth, prediction with the initialized model, and prediction with the trained model. The black dashed lines in the solution field of the ground truth subplot indicate the training data. (B) Loss function. (C) Training data and inferred solution profiles at times $t$= 0.1, 0.3, 0.5, 0.7, and 0.9 s. (D) Unseen ground truth data and inferred solution profiles at selected times $t$= 0.2, 0.4, 0.6, 0.8, and 1.0 s.}
	\label{fig:burger}
\end{figure*}

\hl{Then, the discretized governing equation was solved with a time step of $\Delta t=0.01$ using the ODE integrator of} \verb|Tsit5| \hl{(Tsitouras 5/4 Runge-Kutta method \mbox{\cite{tsitouras_rungekutta_2011}})}. The resultant solution field is illustrated in Fig.~\ref{fig:burger}(A). The solution profiles at the selected times ($t\in\{$0.1, 0.3, 0.5, 0.7, 0.9$\}$) were collected as training data, with an emphasis here on a sparse training dataset that tests the extrapolation capability of the KAN-ODE. To learn the latent solution, we modeled the KAN-ODEs such that

\begin{equation} \label{eq:KAN-ODEs-surrogate}
   \frac{\partial \mathbf{u}\left(t\right)}{\partial t} = \text{KAN}(\mathbf{u}\left(t\right), \bm{\theta}).
\end{equation}

\noindent In this formulation, the \text{KAN} works as a non-linear operator for the partial differential equation (PDE) solution. The KAN-ODE was constructed with 2 layers: [51, 10, 5], [10, 51, 5]. Ten nodes in the intermediate layer were used to preserve the high-dimensional information. Note that the number of input and output dimensions of the KAN were determined by the number of discretized state variables in $\mathbf{u}$. The KAN parameters were updated with a learning rate of 0.01 with the ADAM optimizer.

After 20,000 training epochs (Fig.~\ref{fig:burger}(B)), the KAN-ODE model can be seen in Fig.~\ref{fig:burger}(A) to predict the temporal evolution of the PDE well despite only receiving five training snapshots. These snapshots and their accurate KAN-ODE dynamical representations are shown in Fig.~\ref{fig:burger}(C), where strong agreement is seen at all spatial locations, even those near the shock wave where the training data spatial sampling is very sparse (in addition to sparse temporal sampling). More importantly, the trained KAN-ODE model is able to extrapolate effectively to infer the solution fields at the unseen testing times in Fig.~\ref{fig:burger}(D), once again with strong performance at the shock. As a whole, this case demonstrates the ability of the KAN-ODE to leverage its temporal grid agnosticism to accurately predict a system's behavior at unseen times, and to extrapolate significant meaning from spatially and temporally sparse datasets. An in-depth discussion of the model accuracy and KAN architecture can be found in Appendix B.

\subsubsection{Schr\"{o}dinger Equation}

The last example demonstrates surrogate modeling with KAN-ODEs in a more involved dynamical system with complex-valued states. The Schr\"{o}dinger equation is a classical field equation used in quantum mechanics.

\begin{equation}\label{eq:schrodinger}
   i\frac{\partial u}{\partial t} +\frac{1}{2} \frac{\partial ^{2}u}{\partial x^{2}} + \lvert u \rvert^{2} u=0,
\end{equation}

\noindent where $u\in \mathbb{C}^{1}$. The time and spatial domains were set to $x\in [-5, 5]$ and $t \in [0, \pi/2]$. The initial condition and boundary conditions were defined as following in Eqs.~\ref{eq:schrodinger_ic}-\ref{eq:schrodinger_bc}. \hl{Then, the governing equations were solved with a time step of $\Delta t=0.01$ and grid size of $\Delta x=0.05$ using the stiff ODE integrator of} \verb|Rodas5| \hl{(A 5th order A-stable stiffly stable Rosenbrock method with a stiff-aware 4th order interpolant \mbox{\cite{di1993rodas5,rackauckas_differentialequationsjl_2017}}).} Note that the problem settings used in this example were identical to those in \cite{raissi_physics-informed_2019}.

\begin{align}
    u\left(x,0\right)&=\text{sech}\left(x\right),\label{eq:schrodinger_ic}\\
    u\left(-5,t\right)&=u\left(5,t\right),\label{eq:schrodinger_bc}\\
    \frac{\partial u}{\partial x}\left(-5,t\right)&=\frac{\partial u}{\partial x}\left(5,t\right) \notag.
\end{align}

The KAN-ODE as a surrogate model (Eq.~\ref{eq:KAN-ODEs-surrogate}) was constructed with 2 layers: [402, 10, 10], [10, 402, 10]. The input dimension was doubled from the dimension of the discretized $u$ (201 to 402) to account for both the real and imaginary values of $u$.

\begin{figure*}[tb]
    \centering
	\includegraphics[width=\linewidth]{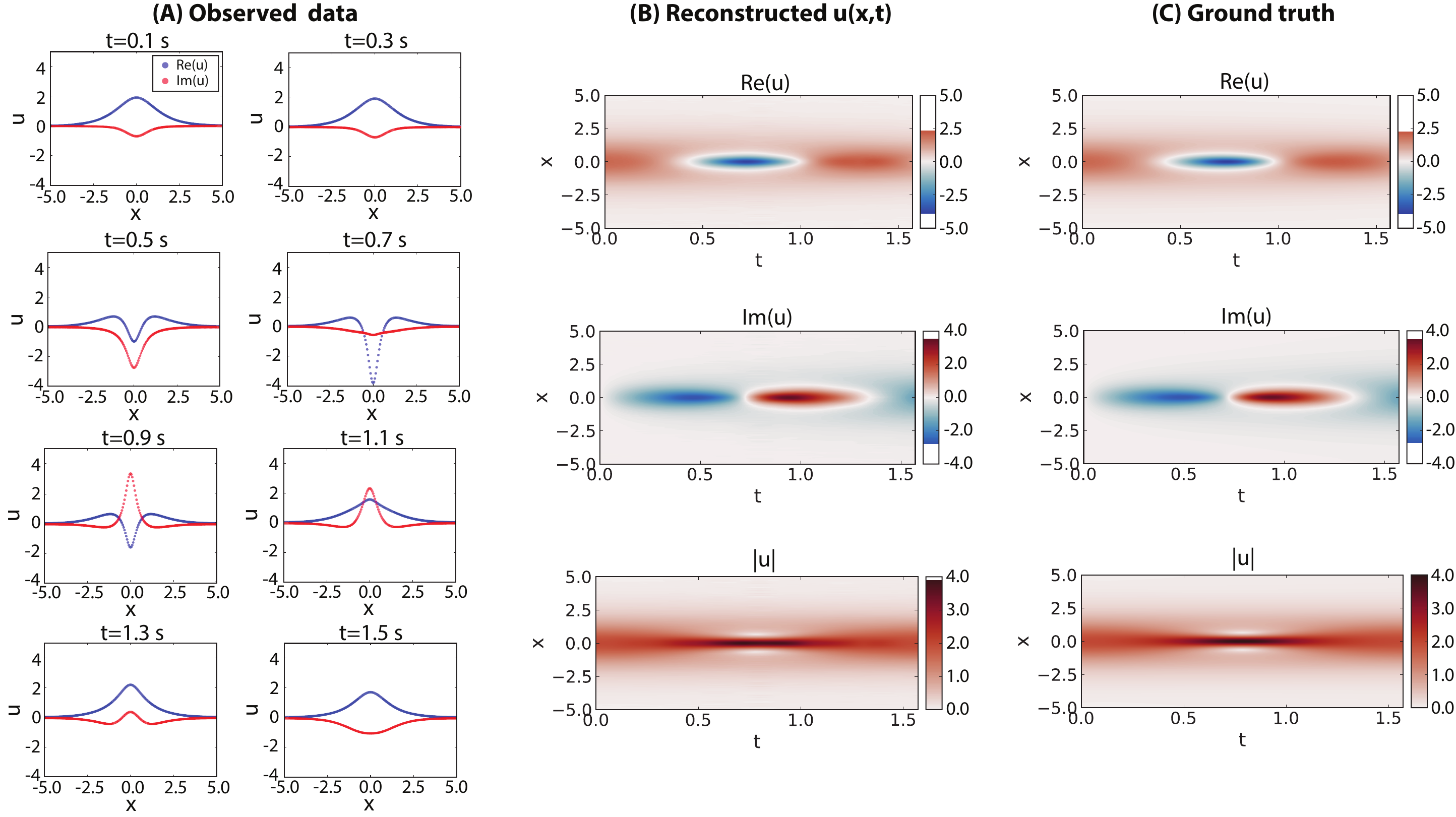}
	\caption{\textit{Schr\"{o}dinger equation}: (A) Observed data $u(x,t)=\text{Re}(u)+i\text{Im}(u)$ at the selected times of $t$ = 0.1, 0.3, 0.5, 0.7, 0.9, 1.1, 1.3, and 1.5 sec as training data. Blue circles denote $\text{Re}(u)$ and red circles denote $\text{Im}(u)$. (B) Reconstructed $u(x,t)$. (C) Ground truth $u(x,t)$. Note that $\text{Re}(u)$ and $\text{Im}(u)$ denote the real and imaginary parts of $u$, respectively. Also, $\lvert u\rvert = \sqrt{\text{Re}(u)^{2} + \text{Im}(u)^{2}}$.}
	\label{fig:schrodinger}
\end{figure*}

We once again gave the KAN-ODE training data at just eight selected times in Fig.~\ref{fig:schrodinger}(A), and yet the trained KAN-ODE successfully reconstructs the complete spatiotemporal profile of the wave field $u(x,t)$ (Fig.~\ref{fig:schrodinger}(B)), which agrees with the ground truth $u(x,t)$ (Fig.~\ref{fig:schrodinger}(C)) at all times. We find this result remarkable, especially as the KAN-ODE had zero knowledge of the underlying physics and instead was forced to infer the complete solution to the nonlinear and complex dynamics present in this system from eight samples only.

In this section, we demonstrated the robustness of KAN-ODEs by learning shock formation behavior and complex-valued wave propagation in data-lean frameworks and without prior knowledge of the system dynamics. In both cases the KAN-ODEs were able to match the training data and generalize well in the significant windows between training data slices, showing promise as a novel data-driven dynamical system modeling approach. With these PDE examples, we showed that KAN-ODEs can retrieve not only temporal dynamics but also spatial information without special treatment in the KAN architecture.

\section{\label{sec:conclusion}Conclusions}

This work proposed the use of Kolmogorov-Arnold networks as gradient-getters in the Neural ODE framework. This novel combination of data-driven techniques advances our capabilities in the field of scientific machine learning to efficiently infer interpretable and modular dynamical system models from various sources of data. The replacement of MLPs with KANs in this framework preserves the black-box style approach with zero prior knowledge of the dynamical system physics needed for training. However, it opens the door to significant interpretability of the learned model ("opening" the black box) via activation function visualization and symbolic regression. Quantitatively, the KAN-ODE framework was also shown to outperform similar MLP-based Neural ODEs in a wide array of metrics including training accuracy, testing accuracy, convergence speed, neural scaling rate, and generalizability. The strong neural scaling offered by the Kolmogorov-Arnold representation held when applied as a KAN-ODE with RBF basis functions, allowing for significantly better training and testing performance in a fraction of the computational time and with a significantly smaller network size. We additionally demonstrated the capability of KAN-ODEs to extrapolate information from sparse or limited temporal data to the rest of a solution domain even in stiff and complex-valued nonlinear PDEs, and to learn symbolic expressions for PDE source terms with high accuracy via optional postprocessing. In addition to these improvements over MLP-based Neural ODEs, KAN-ODEs offer an alternative to SINDy, PINN, and similar interpretable machine learning techniques in their flexibility and generality - no prior knowledge or assumptions are needed about the governing laws or candidate function space to train the KAN-ODE system effectively, although such information can be added if desired as shown here in the source term regression of the Fisher-KPP PDE and the symbolic KAN leveraged for Lotka-Volterra dynamics. KAN-ODEs are a compact, efficient, flexible, and interpretable approach for data-driven modeling of dynamical systems that show significant promise for use in the field of scientific machine learning.

\section*{Data and Materials Availability}

\hl{The KAN-ODE codes used to generate the results presented in this work are publicly available at} \url{https://github.com/DENG-MIT/KAN-ODEs}.

\section*{Acknowledgement}

This work is supported by the National Science Foundation (NSF) under Grant No. CBET-2143625. BCK is partially supported by the NSF Graduate Research Fellowship under Grant No. 1745302. 

\section*{CRediT authorship contribution statement}
\textbf{Benjamin C. Koenig:} Methodology, Software, Investigation, Writing - Original Draft. \textbf{Suyong Kim:} Conceptualization,  Methodology, Software, Investigation, Writing - Original Draft. \textbf{Sili Deng:} Funding Acquisition, Resources, Writing - Review $\&$ Editing.

\section*{Declaration of competing interest}

The authors declare that they have no known competing financial interests or personal relationships that could have appeared to influence the work reported in this paper.

\renewcommand\thetable{A\arabic{table}}
\renewcommand\thefigure{A\arabic{figure}}
\renewcommand\theequation{A\arabic{equation}}
\setcounter{equation}{0} 

\setcounter{figure}{0} 
\setcounter{table}{0}
\setcounter{section}{0} 

\section*{Appendix A. Sparse Regression on Lotka-Volterra KAN-ODE}

Symbolic functions corresponding to the Symbolic KAN-ODE plotted in Fig. \ref{fig:LV2}(B) are provided in Table~\ref{table_LV}.  See Fig.~\ref{fig:LV2}(B) for a visual representation of the twelve activations and of the hidden layer nodes $n_1$ through $n_3$. The selection of candidate functions was carried out to balance accuracy with sparsity. For example, a linear term suffices to capture the linear behavior of $l1$ connecting $x$ and $n_{1}$, while $l2$ connecting $x$ and $n_{2}$ requires an additional $1/x$ term due to its strong nonlinear behavior. Similarly, the relatively complex $l11$ connecting $n_{3}$ to $dx/dt$ uses a linear term in addition to a polynomial division term. All activations were derived independently of each other, based on the immediately relevant univariate input/output pairs.

\begin{table*}[tb] 
\caption{Symbolic KAN activations for Lotka-Volterra dynamics. The first layer takes the inputs $x$ and $y$ while the second layer takes the inputs $n_{1}$, $n_{2}$, and $n_{3}$. Note that $n_{i}(x,y)=n_{i}(x)+n_{i}(y)$ and $dx/dt(n_{1},n_{2},n_{3}) = dx/dt(n_{1})+dx/dt(n_{2})+dx/dt(n_{3})$ by definition in the KAN.}\label{table:symbolic_LV}
 	\begin{ruledtabular}

\begin{tabular}{m{0.1\linewidth}m{0.2\linewidth}m{0.3\linewidth}}
Layer& Activation & Symbolic Expression  \\
\midrule
    1&$\displaystyle n_{1}\left(x\right)$         &      $\displaystyle0.545x-0.204$  \vspace{0.1cm} \\
     &$\displaystyle n_{2}\left(x\right)$   & $\displaystyle -0.277x+0.425+\frac{0.794}{x} $  \vspace{0.1cm} \\
     &$\displaystyle n_{3}\left(x\right)$   &         $\displaystyle0.334x-0.635$  \vspace{0.1cm} \\
      &$\displaystyle n_{1}\left(y\right)$   &    $\displaystyle-0.605y+0.220-\frac{0.120}{y}$   \vspace{0.1cm} \\
      &$\displaystyle n_{2}\left(y\right)$ &          $\displaystyle -0.506y+1.864-\frac{0.108}{y}$   \vspace{0.1cm} \\
       &$\displaystyle n_{3}\left(y\right)$ &          $\displaystyle-0.780y-0.102$  \vspace{0.1cm} \\
\midrule
      2&$\displaystyle \frac{dx}{dt}\left({n_{1}}\right)$  & $\displaystyle0.090n_1^3+0.520n_1^2+0.890n_1-0.411$  \vspace{0.15cm}\\
      &$\displaystyle \frac{dy}{dt}\left({n_1}\right)$     & $\displaystyle0.168n_1^3-0.887n_1^2+2.242n_1+0.748$  \vspace{0.15cm}\\
&$\displaystyle \frac{dx}{dt}\left({n_{2}}\right)$   & $\displaystyle-{n_2^2}+4.990n_2+\frac{0.439n_2}{n_2^2+0.0695}-1.652$  \vspace{0.1cm}\\
&$\displaystyle \frac{dy}{dt}\left({n_2}\right)$  &  $\displaystyle-n_2+\frac{-3.201n_2}{n_2^2+0.719n_2+0.675}$  \vspace{0.1cm}\\
&$\displaystyle \frac{dx}{dt}\left({n_{3}}\right)$  &  $\displaystyle 0.684n_3+\frac{2.929n_3}{n_3^2+1.307n_3+1.41}-0.490$       \vspace{0.1cm}\\
 &$\displaystyle \frac{dy}{dt}\left({n_3}\right)$   &        $\displaystyle-\frac{1.281n_3}{n_3^2+0.800}+0.353$   \vspace{0.1cm}\\
                   
\end{tabular}
       \end{ruledtabular} 

\label{table_LV}
\end{table*}

\section*{Appendix B. Varying KAN-ODE Architectures for Fisher-KPP and Burgers' Equations} \label{sec:appendixB}
\renewcommand\thetable{B\arabic{table}}
\renewcommand\thefigure{B\arabic{figure}}
\renewcommand\theequation{B\arabic{equation}}
\setcounter{equation}{0} 

\setcounter{figure}{0} 
\setcounter{table}{0}
\setcounter{section}{0} 
Additional tests were performed to provide KAN-ODE performance metrics for varying KAN architectures in the \textit{Fisher-KPP equation} and \textit{Burgers' equation} examples. Table~\ref{tab:param_study} shows the test matrix with different sizes of KANs. For the \textit{Fisher-KPP equation}, a relatively small KAN with very few parameters was used to model the hidden physics in the interpretable form (i.e. replacing only the source term). Larger KANs with more parameters (thanks to the larger input/output dimension) were constructed for the \textit{Burgers' equation} to build a surrogate model that predicted the entire solution field. Figure~\ref{fig:architecture_test} illustrates the training losses with respect to the number of parameters in these two examples. We surprisingly found in both cases that small architectures of  KANs, i.e. those with zero to one hidden node and three to five grid points, can successfully learn these dynamical systems with a training loss smaller than 10$^{-6}$. Furthermore, in most cases, KAN-ODEs have a scaling rate of loss $\mathcal{L} \propto N^{-4}$, where $N$ is the total number of parameters in the KAN. This result empirically supports the validity of the scaling law in the original KAN \cite{liu_kan_2024}, as well as results seen in the Lotka-Volterra example shown in Fig. \ref{fig:LV}(C).

\begin{table*}[tb]
	\caption{\label{tab:param_study} KAN-ODE architectures tested for performance evaluation with the Fisher-KPP equation and the Burgers' equation.}
 	\begin{ruledtabular}

\begin{tabular}{lcccc}
	    & Depth & Layer width & Grid size & No. Params \\
     	\hline
	    Fisher-KPP equation & 1 & 1  & [2, 3, 5, 10, 20] & [3, 4, 6, 11, 21]\\ 
	     & 2 & 3  & [2, 3, 5, 10] & [18, 24, 36, 66]\\
     	\hline
	    Burgers' equation & 2 & 1  & [2, 3, 5, 10] & [246, 328, 492, 902] \\ 
	     & 2 & 3  & [2, 3, 5, 10] & [738, 984, 1476, 2706] \\ 
       & 2 & 5  & [2, 3, 5, 10] & [1230, 1640, 2460, 4510] \\ 
        & 2 & 10  & [2, 3, 5, 10] & [2460, 3280, 4920, 9020]\\ 
         & 2 & 20  & [2, 3, 5, 10] & [4920, 6560, 9840, 18040] \\ 
\end{tabular}
\end{ruledtabular}
\end{table*}

\begin{figure*}[tb]
    \centering
    \includegraphics[width=0.7\linewidth]{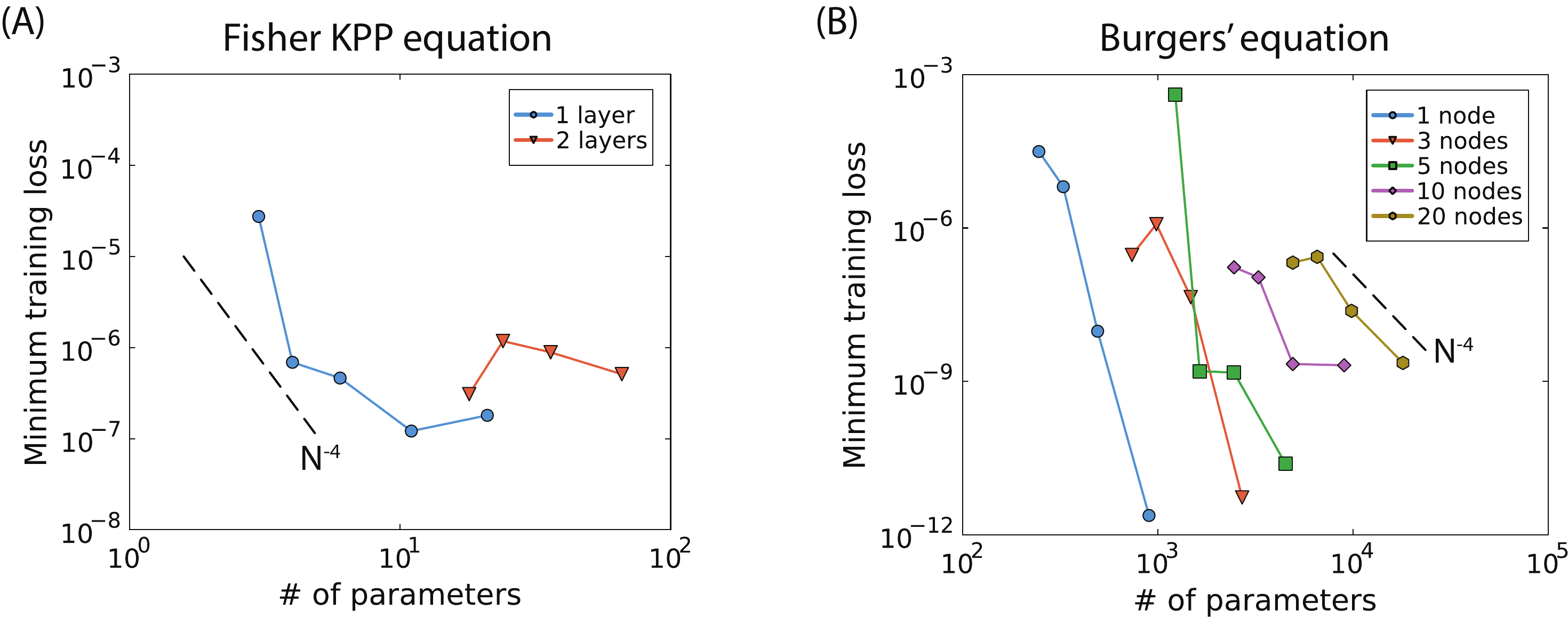}
	\caption{\textit{Performance test with different KAN-ODE architectures}: Training loss as a function of the total number of parameters. (A) Fisher-KPP equation. (B) Burgers' equation. Loss scaling of $\mathcal{L} \propto N^{-4}$, where $N$ is the total number of parameters in the KAN, is also illustrated as a reference line.}
	\label{fig:architecture_test}
\end{figure*}

\renewcommand\thetable{C\arabic{table}}
\renewcommand\thefigure{C\arabic{figure}}
\renewcommand\theequation{C\arabic{equation}}
\setcounter{equation}{0} 
\setcounter{figure}{0} 
\setcounter{table}{0}
\setcounter{section}{0}

\section*{Appendix C. Additional Example with the Allen-Cahn Equation}

In this appendix, we investigated the Allen-Cahn equation which describes phase separation in multi-component alloy systems. The Allen-Cahn equation was used as another canonical example to demonstrate two KAN-ODE tasks: modeling hidden physics and reconstructing complete solution fields. The Allen-Cahn equation is expressed by Eq.~\ref{eq:allen-cahn}.

\begin{equation} \label{eq:allen-cahn}
	\frac{\partial u}{\partial t} = 0.0001 \frac{\partial^2 u}{\partial x^2} + 5u-5u^{3}.
\end{equation}

\noindent The computational domain was defined as $x\in[-1,1]$ and $t\in [0,1]$. The initial and boundary conditions were set as Eqs.~\ref{eq:allen_ic}-\ref{eq:allen_bc}.

\begin{align}
	u\left(x,0\right)&=x^{2} \cos \left(\pi x\right),\label{eq:allen_ic}\\
	u\left(-1,t\right)&=u\left(1,t\right),\label{eq:allen_bc}\\
	\frac{\partial u}{\partial x}\left(-1,t\right)&=\frac{\partial u}{\partial x}\left(1,t\right) \notag.
\end{align}

\noindent Equation~\ref{eq:allen-cahn} was discretized with $\Delta x$ = 0.05 using a central difference scheme. Then, the discretized PDE was solved by the \verb|Tsit5| ODE integrator.

\subsection*{C.1. Modeling Hidden Physics in PDEs}

To model the unknown source term in Eq.~\ref{eq:allen-cahn}, we formulated the KAN-ODE such that

\begin{equation}\label{eq:allen-KAN}
	\frac{\partial u}{\partial t} = 0.0001 \frac{\partial^2 u}{\partial x^2} + \text{KAN}\left(u, \bm{\theta}\right). 
\end{equation}

\noindent Similarly to the \textit{Fisher-KPP} example, Eq.~\ref{eq:allen-KAN} was discretized using a central difference scheme and converted to an ODE system using the Method of Lines. We constructed the KAN model with a single layer comprising a single node, [1,1,10], thereby encoding just one activation function.

\begin{figure*}[tb]
    \centering
	\includegraphics[width=\linewidth]{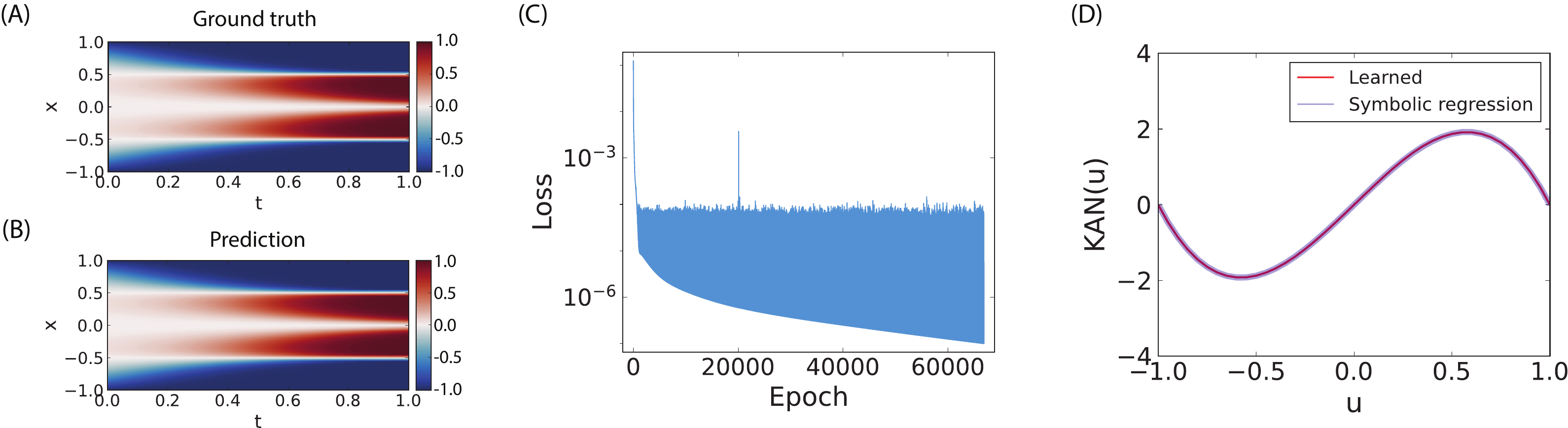}
	\caption{\textit{Allen-Cahn equation}: (A) Solution field $u(x,t)$ of the ground truth. (B) Solution field $u(x,t)$ of the prediction with a trained KAN-ODE model. (C) Loss function. (D) Learned hidden physics of the reaction source term in the Allen-Cahn equation, and its symbolic expression.}
	\label{fig:allen_source}
\end{figure*}

The KAN-ODE model was trained on the data shown in Fig.~\ref{fig:allen_source}(A). In this example, we applied an early stop when the error reached 10$^{-7}$. The prediction with the learned model is illustrated in Fig.~\ref{fig:allen_source}(B), showing accurate reconstruction of the solution field $u(x,t)$ after $\sim$65,000 updates (Fig.~\ref{fig:allen_source}(C)). This resultant single activation function is shown in Fig.~\ref{fig:allen_source}(D). For the modeled activation function, a symbolic expression was derived, using the SymbolicRegression.jl package \cite{cranmerInterpretableMachineLearning2023}. Given the candidate symbolic expressions of $[+,-,\times,/]$, it returned the expressions tabulated in Table~\ref{tab:allen_symbolic}. Among them, the following was the simplest expression of the low-loss group (i.e. the group of expressions that jumped from $10^{-1}$ to $10^{-5}$ loss),

\begin{table*}[tb]
	\caption{\label{tab:allen_symbolic} Seven candidate symbolic expressions of the reaction source term in the Allen-Cahn equation (Eq.~\ref{eq:allen-cahn}). The complexities of the symbolic equations and the corresponding losses are also tabulated.}
 \begin{ruledtabular}

\begin{tabular}{ccl}
			Complexity & Loss & Equation \\
			\midrule
			1 & 9.1219$\times$10$^{-1}$ &  $u$ \\ 
			3 & 6.5775$\times$10$^{-1}$ &  $1.8526u$ \\ 
			5 & 6.5775$\times$10$^{-1}$ &  $u/0.5398+8.3105\times10^{-4}$ \\ 
			7 & 4.2106$\times$10$^{-1}$ &  $(2.4821-u^{2})u$ \\ 
			9 & 1.1523$\times$10$^{-5}$ &  $(5.0015-5.0021u^{2})u$ \\ 
			11 & 1.0832$\times$10$^{-5}$ &  $5.0021(0.9999-u^{2})u$ \\ 
			13 & 1.0493$\times$10$^{-5}$ &  $5.0021u(0.9999-u(u+3.7283\times 10^{-4}))$ \\ 
		\end{tabular}
	\end{ruledtabular}
\end{table*}

\begin{equation}
	\text{KAN}\left(u\right) = (5.0015-5.0021u^{2})u \label{eq:allen_source_symbol}.
\end{equation}

\noindent Equation~\ref{eq:allen_source_symbol} is remarkably close to the exact source term of $5u-5u^{3}$. Note that all the symbolic expressions with errors on the order of 10$^{-5}$ in Table~\ref{tab:allen_symbolic} have an identical form with slightly different coefficients. This example provides another demonstration of accurate and human-interpretable modeling of hidden physics from measurable quantities via KAN-ODEs.

\subsection*{C.2. Data-Driven Solutions of PDEs}

\begin{figure*}[tb]
    \centering
	\includegraphics[width=0.65\linewidth]{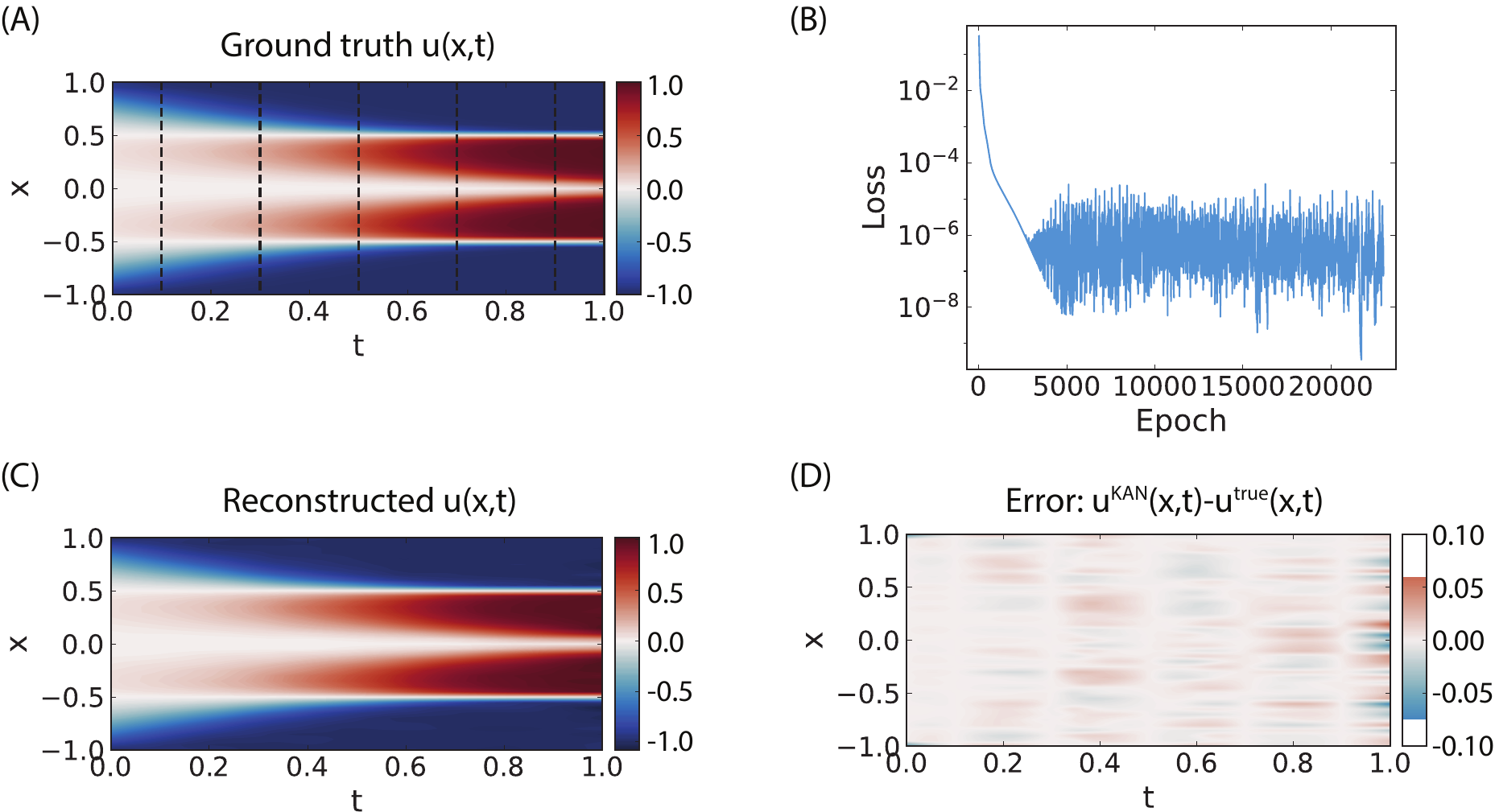}
	\caption{\textit{Allen-Cahn equation}: Demonstration of a KAN-ODE as a surrogate model. (A) Ground truth $u(x,t)$. The dashed lines indicate the training data at $t$ = 0.1, 0.3, 0.5, 0.7, 0.9. (B) MSE loss during training. (C) Reconstructed $u(x,t)$ by the trained KAN-ODE model. (D) The error between the reconstructed $u^{\text{KAN}}(x,t)$ and the ground truth $u^{\text{true}}(x,t)$.}
 	\label{fig:allen_surrogate}
\end{figure*}

We further demonstrate KAN-ODEs as surrogate models using the Allen-Cahn equation by replacing the entire right-hand side of Eq. \ref{eq:allen-cahn} with a KAN. To mimic sparse data measurements, only five profiles at the selected times $t\in\{0.1, 0.3, 0.5, 0.7, 0.9\}$ were collected from the ground truth simulation and used as training data (Fig.~\ref{fig:allen_surrogate}(A)), while testing involved reconstruction of the entire time history. The KAN-ODE as a surrogate model was constructed using 2 layers with 10 hidden nodes ([41,10,10], [10,41,10]). Equation~\ref{eq:KAN-ODEs-surrogate} was trained to learn the dynamics of the phase transition as shown in the loss plot (Fig.~\ref{fig:allen_surrogate}(B)). The resulting field reconstruction of $u(x,t)$ is illustrated in Fig.~\ref{fig:allen_surrogate}(C), which qualitatively matches well with the ground truth. The spatio-temporal error between the KAN-ODE model and the ground truth ($u^{\text{KAN}}(x,t)-u^{\text{true}}(x,t)$) in Fig.~\ref{fig:allen_surrogate}(D) depicts near-zero error at and surrounding each training time slice, and small but noticeable error in the large windows between time steps, especially near the end of the simulation time. For example, the error is not appreciable at time $t$ = 0.1, but begins to grow before shrinking back to near-zero by $t$ = 0.3 when the second observation takes place. While the errors introduced here are fairly minor, this result implies that the KAN-ODE needs more training data in order to extrapolate quantitatively well to times significantly outside of its training data window, like $t$ = 1.5 s or $t$ = 2.0 s. Within the 1 s range considered here, however, these results highlight the strong capability of KAN-ODEs to accurately model and reconstruct physical phenomena between training samples with very sparse measurements and \textit{without any prior physical knowledge} included. 

\bibliographystyle{elsarticle-num}
\bibliography{KANODE}

\begin{thebibliography}{10}
\expandafter\ifx\csname url\endcsname\relax
  \def\url#1{\texttt{#1}}\fi
\expandafter\ifx\csname urlprefix\endcsname\relax\def\urlprefix{URL }\fi
\expandafter\ifx\csname href\endcsname\relax
  \def\href#1#2{#2} \def\path#1{#1}\fi

\bibitem{karniadakis2021physics}
G.~E. Karniadakis, I.~G. Kevrekidis, L.~Lu, P.~Perdikaris, S.~Wang, L.~Yang, Physics-informed machine learning, Nature Reviews Physics 3~(6) (2021) 422--440.
\newblock \href {https://doi.org/10.1038/s42254-021-00314-5} {\path{doi:10.1038/s42254-021-00314-5}}.

\bibitem{brunton2016discovering}
S.~L. Brunton, J.~L. Proctor, J.~N. Kutz, {Discovering governing equations from data by sparse identification of nonlinear dynamical systems}, Proceedings of the National Academy of Sciences 113~(15) (2016) 3932--3937.
\newblock \href {https://doi.org/10.1073/pnas.1517384113} {\path{doi:10.1073/pnas.1517384113}}.

\bibitem{brunton2024promising}
S.~L. Brunton, J.~N. Kutz, Promising directions of machine learning for partial differential equations, Nature Computational Science (2024) 1--12\href {https://doi.org/10.1038/s43588-024-00643-2} {\path{doi:10.1038/s43588-024-00643-2}}.

\bibitem{rudy_data-driven_2017}
S.~H. Rudy, S.~L. Brunton, J.~L. Proctor, J.~N. Kutz, Data-driven discovery of partial differential equations, Science Advances 3~(4) (2017) e1602614.
\newblock \href {https://doi.org/10.1126/sciadv.1602614} {\path{doi:10.1126/sciadv.1602614}}.

\bibitem{brunton_discovering_2016}
S.~L. Brunton, J.~L. Proctor, J.~N. Kutz, Discovering governing equations from data by sparse identification of nonlinear dynamical systems, Proceedings of the National Academy of Sciences 113~(15) (2016) 3932--3937.
\newblock \href {https://doi.org/10.1073/pnas.1517384113} {\path{doi:10.1073/pnas.1517384113}}.

\bibitem{raissi_physics-informed_2019}
M.~Raissi, P.~Perdikaris, G.~E. Karniadakis, Physics-informed neural networks: {A} deep learning framework for solving forward and inverse problems involving nonlinear partial differential equations, Journal of Computational Physics 378 (2019) 686--707.
\newblock \href {https://doi.org/10.1016/j.jcp.2018.10.045} {\path{doi:10.1016/j.jcp.2018.10.045}}.

\bibitem{chen_neural_2019}
R.~T.~Q. Chen, Y.~Rubanova, J.~Bettencourt, D.~Duvenaud, Neural {Ordinary} {Differential} {Equations}, arXiv:1806.07366 (2019).
\newblock \href {https://doi.org/10.48550/arXiv.1806.07366} {\path{doi:10.48550/arXiv.1806.07366}}.

\bibitem{kim_stiff_2021}
S.~Kim, W.~Ji, S.~Deng, Y.~Ma, C.~Rackauckas, Stiff {Neural} {Ordinary} {Differential} {Equations}, Chaos: An Interdisciplinary Journal of Nonlinear Science 31~(9) (2021) 093122.
\newblock \href {https://doi.org/10.1063/5.0060697} {\path{doi:10.1063/5.0060697}}.

\bibitem{dandekar_bayesian_2022}
R.~Dandekar, K.~Chung, V.~Dixit, M.~Tarek, A.~Garcia-Valadez, K.~V. Vemula, C.~Rackauckas, Bayesian {Neural} {Ordinary} {Differential} {Equations}, arXiv:2012.07244 (2022).
\newblock \href {https://doi.org/10.48550/arXiv.2012.07244} {\path{doi:10.48550/arXiv.2012.07244}}.

\bibitem{lu2018beyond}
Y.~Lu, A.~Zhong, Q.~Li, B.~Dong, Beyond finite layer neural networks: Bridging deep architectures and numerical differential equations, in: International Conference on Machine Learning, PMLR, 2018, pp. 3276--3285.

\bibitem{haber2017stable}
E.~Haber, L.~Ruthotto, Stable architectures for deep neural networks, Inverse problems 34~(1) (2017) 014004.
\newblock \href {https://doi.org/10.1088/1361-6420/aa9a90} {\path{doi:10.1088/1361-6420/aa9a90}}.

\bibitem{ji_stiff-pinn_2021}
W.~Ji, W.~Qiu, Z.~Shi, S.~Pan, S.~Deng, Stiff-{PINN}: {Physics}-{Informed} {Neural} {Network} for {Stiff} {Chemical} {Kinetics}, The Journal of Physical Chemistry A 125~(36) (2021) 8098--8106.
\newblock \href {https://doi.org/10.1021/acs.jpca.1c05102} {\path{doi:10.1021/acs.jpca.1c05102}}.

\bibitem{cuomo_scientific_2022}
S.~Cuomo, V.~S. Di~Cola, F.~Giampaolo, G.~Rozza, M.~Raissi, F.~Piccialli, Scientific {Machine} {Learning} {Through} {Physics}–{Informed} {Neural} {Networks}: {Where} we are and {What}’s {Next}, Journal of Scientific Computing 92~(3) (2022) 88.
\newblock \href {https://doi.org/10.1007/s10915-022-01939-z} {\path{doi:10.1007/s10915-022-01939-z}}.

\bibitem{geneva_modeling_2020}
N.~Geneva, N.~Zabaras, Modeling the dynamics of {PDE} systems with physics-constrained deep auto-regressive networks, Journal of Computational Physics 403 (2020) 109056.
\newblock \href {https://doi.org/10.1016/j.jcp.2019.109056} {\path{doi:10.1016/j.jcp.2019.109056}}.

\bibitem{ren_phycrnet_2022}
P.~Ren, C.~Rao, Y.~Liu, J.-X. Wang, H.~Sun, {PhyCRNet}: {Physics}-informed convolutional-recurrent network for solving spatiotemporal {PDEs}, Computer Methods in Applied Mechanics and Engineering 389 (2022) 114399.
\newblock \href {https://doi.org/10.1016/j.cma.2021.114399} {\path{doi:10.1016/j.cma.2021.114399}}.

\bibitem{lu_physics-informed_2021}
L.~Lu, R.~Pestourie, W.~Yao, Z.~Wang, F.~Verdugo, S.~G. Johnson, Physics-{Informed} {Neural} {Networks} with {Hard} {Constraints} for {Inverse} {Design}, SIAM Journal on Scientific Computing 43~(6) (2021) B1105--B1132.
\newblock \href {https://doi.org/10.1137/21M1397908} {\path{doi:10.1137/21M1397908}}.

\bibitem{alkhadhr_wave_2023}
S.~Alkhadhr, M.~Almekkawy, Wave {Equation} {Modeling} via {Physics}-{Informed} {Neural} {Networks}: {Models} of {Soft} and {Hard} {Constraints} for {Initial} and {Boundary} {Conditions}, Sensors 23~(5) (2023) 2792.
\newblock \href {https://doi.org/10.3390/s23052792} {\path{doi:10.3390/s23052792}}.

\bibitem{lai_structural_2021}
Z.~Lai, C.~Mylonas, S.~Nagarajaiah, E.~Chatzi, Structural identification with physics-informed neural ordinary differential equations, Journal of Sound and Vibration 508 (2021) 116196.
\newblock \href {https://doi.org/10.1016/j.jsv.2021.116196} {\path{doi:10.1016/j.jsv.2021.116196}}.

\bibitem{ji_autonomous_2021}
W.~Ji, S.~Deng, Autonomous {Discovery} of {Unknown} {Reaction} {Pathways} from {Data} by {Chemical} {Reaction} {Neural} {Network}, The Journal of Physical Chemistry A 125~(4) (2021) 1082--1092.
\newblock \href {https://doi.org/10.1021/acs.jpca.0c09316} {\path{doi:10.1021/acs.jpca.0c09316}}.

\bibitem{ji_autonomous_2022}
W.~Ji, F.~Richter, M.~J. Gollner, S.~Deng, Autonomous kinetic modeling of biomass pyrolysis using chemical reaction neural networks, Combustion and Flame 240 (2022) 111992.
\newblock \href {https://doi.org/10.1016/j.combustflame.2022.111992} {\path{doi:10.1016/j.combustflame.2022.111992}}.

\bibitem{koenig_accommodating_2023}
B.~C. Koenig, P.~Zhao, S.~Deng, Accommodating physical reaction schemes in {DSC} cathode thermal stability analysis using chemical reaction neural networks, Journal of Power Sources 581 (2023) 233443.
\newblock \href {https://doi.org/10.1016/j.jpowsour.2023.233443} {\path{doi:10.1016/j.jpowsour.2023.233443}}.

\bibitem{li_bayesian_2023}
Q.~Li, H.~Chen, B.~C. Koenig, S.~Deng, Bayesian chemical reaction neural network for autonomous kinetic uncertainty quantification, Physical Chemistry Chemical Physics 25~(5) (2023) 3707--3717.
\newblock \href {https://doi.org/10.1039/D2CP05083H} {\path{doi:10.1039/D2CP05083H}}.

\bibitem{koenig_uncertain_2024}
B.~C. Koenig, H.~Chen, Q.~Li, P.~Zhao, S.~Deng, Uncertain lithium-ion cathode kinetic decomposition modeling via {Bayesian} chemical reaction neural networks, Proceedings of the Combustion Institute 40~(1) (2024) 105243.
\newblock \href {https://doi.org/10.1016/j.proci.2024.105243} {\path{doi:10.1016/j.proci.2024.105243}}.

\bibitem{kaiser_sparse_2018}
E.~Kaiser, J.~N. Kutz, S.~L. Brunton, Sparse identification of nonlinear dynamics for model predictive control in the low-data limit, Proceedings of the Royal Society A: Mathematical, Physical and Engineering Sciences 474~(2219) (2018) 20180335.
\newblock \href {https://doi.org/10.1098/rspa.2018.0335} {\path{doi:10.1098/rspa.2018.0335}}.

\bibitem{fasel_ensemble-sindy_2022}
U.~Fasel, J.~N. Kutz, B.~W. Brunton, S.~L. Brunton, Ensemble-{SINDy}: {Robust} sparse model discovery in the low-data, high-noise limit, with active learning and control, Proceedings of the Royal Society A: Mathematical, Physical and Engineering Sciences 478~(2260) (2022) 20210904, arXiv:2111.10992.
\newblock \href {https://doi.org/10.1098/rspa.2021.0904} {\path{doi:10.1098/rspa.2021.0904}}.

\bibitem{liu_kan_2024}
Z.~Liu, Y.~Wang, S.~Vaidya, F.~Ruehle, J.~Halverson, M.~Soljačić, T.~Y. Hou, M.~Tegmark, {KAN}: {Kolmogorov}-{Arnold} {Networks}, arXiv:2404.19756 (2024).
\newblock \href {https://doi.org/10.48550/arXiv.2404.19756} {\path{doi:10.48550/arXiv.2404.19756}}.

\bibitem{vaca-rubio_kolmogorov-arnold_2024}
C.~J. Vaca-Rubio, L.~Blanco, R.~Pereira, M.~Caus, Kolmogorov-{Arnold} {Networks} ({KANs}) for {Time} {Series} {Analysis}, arXiv:2405.08790 (2024).
\newblock \href {https://doi.org/10.48550/arXiv.2405.08790} {\path{doi:10.48550/arXiv.2405.08790}}.

\bibitem{xu_kolmogorov-arnold_2024}
K.~Xu, L.~Chen, S.~Wang, Kolmogorov-{Arnold} {Networks} for {Time} {Series}: {Bridging} {Predictive} {Power} and {Interpretability}, arXiv:2406.02496 (2024).
\newblock \href {https://doi.org/10.48550/arXiv.2406.02496} {\path{doi:10.48550/arXiv.2406.02496}}.

\bibitem{hornik_multilayer_1989}
K.~Hornik, M.~Stinchcombe, H.~White, Multilayer feedforward networks are universal approximators, Neural Networks 2~(5) (1989) 359--366.
\newblock \href {https://doi.org/10.1016/0893-6080(89)90020-8} {\path{doi:10.1016/0893-6080(89)90020-8}}.

\bibitem{kolmogorov_representation_1956}
A.~N. Kolmogorov, On the representation of continuous functions of several variables as superpositions of continuous functions of a smaller number of variables., Dokl. Akad. Nauk~(108(2)) (1956).

\bibitem{li_kolmogorov-arnold_2024}
Z.~Li, Kolmogorov-{Arnold} {Networks} are {Radial} {Basis} {Function} {Networks}, arXiv:2405.06721 (2024).
\newblock \href {https://doi.org/10.48550/arXiv.2405.06721} {\path{doi:10.48550/arXiv.2405.06721}}.

\bibitem{puri_kolmogorovarnoldjl_2024}
V.~Puri, {KolmogorovArnold}.jl, {GitHub repository. Retrieved from https://github.com/vpuri3/KolmogorovArnold.jl}. (2024).

\bibitem{ramachandran_searching_2017}
P.~Ramachandran, B.~Zoph, Q.~V. Le, Searching for {Activation} {Functions}, arXiv:1710.05941 (2017).
\newblock \href {https://doi.org/10.48550/arXiv.1710.05941} {\path{doi:10.48550/arXiv.1710.05941}}.

\bibitem{kim2023inference}
S.~Kim, S.~Deng, {Inference of chemical kinetics and thermodynamic properties from constant-volume combustion of energetic materials}, Chemical Engineering Journal 469 (2023) 143779.
\newblock \href {https://doi.org/10.1016/j.cej.2023.143779} {\path{doi:10.1016/j.cej.2023.143779}}.

\bibitem{rackauckas2020universal}
C.~Rackauckas, Y.~Ma, J.~Martensen, C.~Warner, K.~Zubov, R.~Supekar, D.~Skinner, A.~Ramadhan, A.~Edelman, Universal {Differential} {Equations} for {Scientific} {Machine} {Learning}, arXiv:2001.04385 (2021).
\newblock \href {https://doi.org/10.48550/arXiv.2001.04385} {\path{doi:10.48550/arXiv.2001.04385}}.

\bibitem{maly1996numerical}
T.~Maly, L.~R. Petzold, Numerical methods and software for sensitivity analysis of differential-algebraic systems, Appl. Numer. Math. 20~(1-2) (1996) 57--79.
\newblock \href {https://doi.org/10.1016/0168-9274(95)00117-4} {\path{doi:10.1016/0168-9274(95)00117-4}}.

\bibitem{cao2002adjoint}
Y.~Cao, S.~Li, L.~Petzold, Adjoint sensitivity analysis for differential-algebraic equations: algorithms and software, J. Comput. Appl. Math. 149~(1) (2002) 171--191.
\newblock \href {https://doi.org/10.1016/S0377-0427(02)00528-9} {\path{doi:10.1016/S0377-0427(02)00528-9}}.

\bibitem{cao2003adjoint}
Y.~Cao, S.~Li, L.~Petzold, R.~Serban, Adjoint sensitivity analysis for differential-algebraic equations: The adjoint dae system and its numerical solution, SIAM J. Sci. Comput. 24~(3) (2003) 1076--1089.
\newblock \href {https://doi.org/10.1137/S1064827501380630} {\path{doi:10.1137/S1064827501380630}}.

\bibitem{rackauckas_differentialequationsjl_2017}
C.~Rackauckas, Q.~Nie, {DifferentialEquations}.jl – {A} {Performant} and {Feature}-{Rich} {Ecosystem} for {Solving} {Differential} {Equations} in {Julia}, Journal of Open Research Software 5~(1) (2017).
\newblock \href {https://doi.org/10.5334/jors.151} {\path{doi:10.5334/jors.151}}.

\bibitem{pal2023lux}
A.~Pal, {Lux: Explicit Parameterization of Deep Neural Networks in Julia} (2023).
\newblock \href {https://doi.org/10.5281/zenodo.7808904} {\path{doi:10.5281/zenodo.7808904}}.

\bibitem{tsitouras_rungekutta_2011}
C.~Tsitouras, Runge–{Kutta} pairs of order 5(4) satisfying only the first column simplifying assumption, Computers \& Mathematics with Applications 62~(2) (2011) 770--775.
\newblock \href {https://doi.org/10.1016/j.camwa.2011.06.002} {\path{doi:10.1016/j.camwa.2011.06.002}}.

\bibitem{kingma_adam_2017}
D.~P. Kingma, J.~Ba, Adam: {A} {Method} for {Stochastic} {Optimization}, arXiv:1412.6980 (2017).
\newblock \href {https://doi.org/10.48550/arXiv.1412.6980} {\path{doi:10.48550/arXiv.1412.6980}}.

\bibitem{michaud_precision_2023}
E.~J. Michaud, Z.~Liu, M.~Tegmark, Precision {Machine} {Learning}, Entropy 25~(1) (2023) 175.
\newblock \href {https://doi.org/10.3390/e25010175} {\path{doi:10.3390/e25010175}}.

\bibitem{blealtan_efficient-kan_2024}
Blealtan, \href{https://github.com/Blealtan/efficient-kan}{efficient-kan} (2024).
\newline\urlprefix\url{https://github.com/Blealtan/efficient-kan}

\bibitem{cranmerInterpretableMachineLearning2023}
M.~Cranmer, Interpretable {Machine} {Learning} for {Science} with {PySR} and {SymbolicRegression}.jl, arXiv:2305.01582 (2023).
\newblock \href {https://doi.org/10.48550/arXiv.2305.01582} {\path{doi:10.48550/arXiv.2305.01582}}.

\bibitem{liu_neural_2019}
X.~Liu, T.~Xiao, S.~Si, Q.~Cao, S.~Kumar, C.-J. Hsieh, Neural {SDE}: {Stabilizing} {Neural} {ODE} {Networks} with {Stochastic} {Noise}, arXiv:1906.02355 (2019).
\newblock \href {https://doi.org/10.48550/arXiv.1906.02355} {\path{doi:10.48550/arXiv.1906.02355}}.

\bibitem{ying_overview_2019}
X.~Ying, \href{https://dx.doi.org/10.1088/1742-6596/1168/2/022022}{An {Overview} of {Overfitting} and its {Solutions}}, Journal of Physics: Conference Series 1168~(2) (2019) 022022.
\newline\urlprefix\url{https://dx.doi.org/10.1088/1742-6596/1168/2/022022}

\bibitem{srivastava_dropout_2014}
N.~Srivastava, G.~Hinton, A.~Krizhevsky, I.~Sutskever, R.~Salakhutdinov, Dropout: {A} {Simple} {Way} to {Prevent} {Neural} {Networks} from {Overfitting}, Journal of Machine Learning Research 15~(56) (2014) 1929--1958.

\bibitem{di1993rodas5}
G.~Di~Marzo, Rodas5 (4)-m{\'e}thodes de rosenbrock d’ordre 5 (4) adapt{\'e}es aux problemes diff{\'e}rentiels-alg{\'e}briques, MSc Mathematics Thesis (1993).

\end{thebibliography}

\end{document}